\documentclass{article}
\usepackage{jfrExamplee}
\usepackage{graphicx}
\usepackage{apalike}
\usepackage{setspace}
\usepackage{enumitem} 
\usepackage{hyperref}
\usepackage{amsmath,amsfonts}
\usepackage{algorithm,algpseudocode}
\usepackage{booktabs}
\usepackage{color}
\usepackage{cite}
\usepackage{amssymb}
\usepackage{subfigure}
\usepackage{array}
\usepackage{easyReview}
\usepackage{acronym}
\usepackage{colortbl}
\usepackage{multirow}
\usepackage{rotating}
\acrodef{IMU}{inertial measurement unit}
\acrodefplural{IMUs}{inertial measurement units}
\acrodef{AUV}{autonomous underwater vehicle}
\acrodefplural{AUVs}{autonomous underwater vehicles}
\acrodef{UAV}{unmanned aerial vehicle}
\acrodefplural{UAVs}{unmanned aerial vehicles}
\acrodef{MAV}{micro aerial vehicle}
\acrodefplural{MAVs}{micro aerial vehicles}
\acrodef{VO}{visual odometry}
\acrodef{VIO}{visual inertial odometry}
\acrodef{SLAM}{Simultaneous Localization and Mapping}
\acrodef{ZUPT}{Zero-Velocity Update}
\acrodefplural{ZUPTs}{Zero Velocity Updates}
\acrodef{ZARU}{Zero Angular Rate Update}
\acrodefplural{ZARUs}{Zero Angular Rate Updates}
\acrodef{TOED}{Theory of Optimal Experiment Design}
\acrodef{TRCTL}{Traction Control}
\acrodef{LQR}{Linear Quadratic Regulator}
\acrodef{CDF}{Cumulative Distribution Function}
\acrodef{RMSE}{Root Mean Square Error}
\acrodef{USGS}{United States Geological Survey}
\acrodef{GNSS}{Global Navigation Satellite System}
\acrodef{GPS}{Global Positioning System}
\acrodef{DGPS}{Differential Global Positioning System}
\acrodef{PPP}{Precise Point Positioning}
\acrodef{GP}{Gaussian Process}
\acrodef{GIS}{Geographic Information System}
\acrodef{TWD}{think-while-driving}
\acrodef{MSL}{Mars Science Laboratory}
\acrodef{MER}{Mars Exploration Rover}
\acrodefplural{MERs}{Mars Exploration Rovers}
\acrodef{MSR}{Mars Sample Return}
\acrodef{INS}{inertial navigation system}
\acrodef{DEM}{digital elevation map}
\acrodef{HiRISE}{High Resolution Imaging Science Experiment}
\acrodef{FPGA}{field-programmable gate array}
\acrodef{WO}{wheel odometry}
\acrodef{WIO}{wheel inertial odometry}
\acrodef{JPL}{Jet Propulsion Laboratory}
\acrodef{BD}{blind driving}
\acrodef{ROS}{Robot Operating System}
\acrodef{EKF}{extended Kalman filter}
\acrodef{STM}{state transition matrix}

\newcolumntype{M}[1]{>{\centering\arraybackslash}m{#1}}

\title{Proprioceptive Slip Detection for Planetary Rovers in Perceptually Degraded Extraterrestrial Environments}

\author{
Cagri Kilic, Yu Gu, and Jason N. Gross  \\
Department of Mechanical and Aerospace Engineering \\
West Virginia University\\
Morgantown, WV 26506 \\

}

\usepackage[pscoord]{eso-pic}
\newcommand{\placetextbox}[3]{
\setbox0=\hbox{#3}
\AddToShipoutPictureFG{ \put(\LenToUnit{#1\paperwidth},\LenToUnit{#2\paperheight}){\vtop{{\null}\makebox[0pt][c]{#3}}}}
}
\placetextbox{.5}{0.05}{Preprint version. Accepted for publication in Field Robotics on 27-July-2022. Submitted on 02-Nov-2021. } 

\begin{document}

\maketitle

\begin{abstract}
Slip detection is of fundamental importance for the safety and efficiency of rovers driving on the surface of extraterrestrial bodies. Current planetary rover slip detection systems rely on visual perception on the assumption that sufficient visual features can be acquired in the environment. However, visual-based methods are prone to suffer in perceptually degraded planetary environments with dominant low terrain features such as regolith, glacial terrain, salt-evaporites, and poor lighting conditions such as dark caves and permanently shadowed regions. Relying only on visual sensors for slip detection also requires additional computational power and reduces the rover traversal rate. This paper answers the question of how to detect wheel slippage of a planetary rover without depending on visual perception. In this respect, we propose a slip detection system that obtains its information from a proprioceptive localization framework that is capable of providing reliable, continuous, and computationally efficient state estimation over hundreds of meters. This is accomplished by using zero velocity update, zero angular rate update, and non-holonomic constraints as pseudo-measurement updates on an inertial navigation system framework. The proposed method is evaluated on actual hardware and field-tested in a planetary-analog environment. The method achieves greater than 92\% slip detection accuracy for distances around 150 m using only an \ac{IMU} and wheel encoders.

\end{abstract}

\section{Introduction}

\subsection{Motivation}

Acquiring accurate slip detection is one of the critical capabilities required for planetary rovers~\cite{heverly2013traverse} to maintain safer driving conditions. Wheel slippage is often unavoidable for a planetary rover, it affects the traction and energy consumption and causes the significant drift from rover's planned path and poor results in the rover state estimates~\cite{c12}. Due to radiation-hardened hardware requirements, slip detection capability is challenging to achieve for the rovers with limited energy sources and computational power. Early phases of planetary rover localization methods have exhibited large onboard localization errors and have had to rely primarily on human-in-the-loop operations \cite{c1}. For example, maintaining a pose estimate with using \ac{IMU} and wheel encoders is achievable on benign terrains; however, due to \ac{INS} drift and wheel slip, \ac{WO}-based localization is often challenging that results in a significant problem for rover localization over time \cite{Maimone07}. For this reason, Mars rovers have been substantially benefited from stereo vision-based odometry to detect slip and compute position updates whereas \ac{IMU} provides the attitude solution\cite{rankin2020driving}. 

Despite their reliability, vision-based systems operate with assuming that the terrain contains sufficient visual texture for localization. This assumption poses a challenge on extraterrestrial bodies that adequate visual features are lacking in the region (e.g., glacier ice, regolith, salt evaporites) \cite{c21} or when the lighting conditions are insufficient~\cite{zhang2015visual}. 

\begin{figure*}[h]
\centerline
{
	\subfigure[Image was taken by Opportunity on the plains of Meridiani at a place known as "Purgatory Dune" during sol 491. Visual feature detection for navigation is often insufficient to find adequate features in the sandy regions. Credit: NASA/JPL-Caltech]
	{
		\includegraphics[width=0.47\columnwidth]{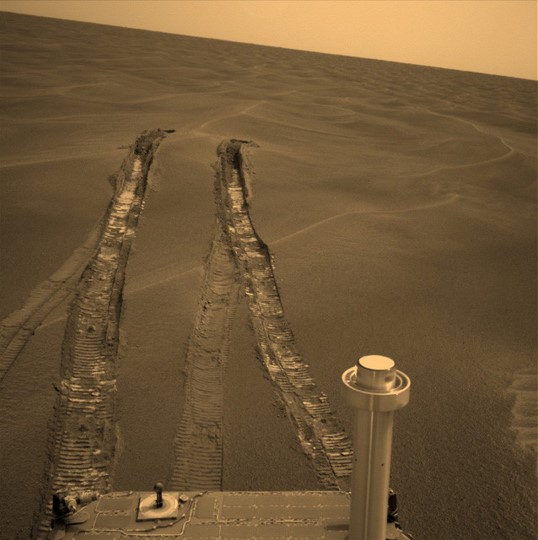}
		\label{fig:purgatory}
	}
	\hspace{0.5\baselineskip}
	\subfigure[This mosaic shows a region on Europa, where the surface has broken apart into many smaller chaos blocks that are surrounded by featureless matrix material. The mosaic images were taken on September 26th, 1998 by NASA's Galileo spacecraft. Credit: NASA   ]
	{
		\includegraphics[width=0.47\columnwidth]{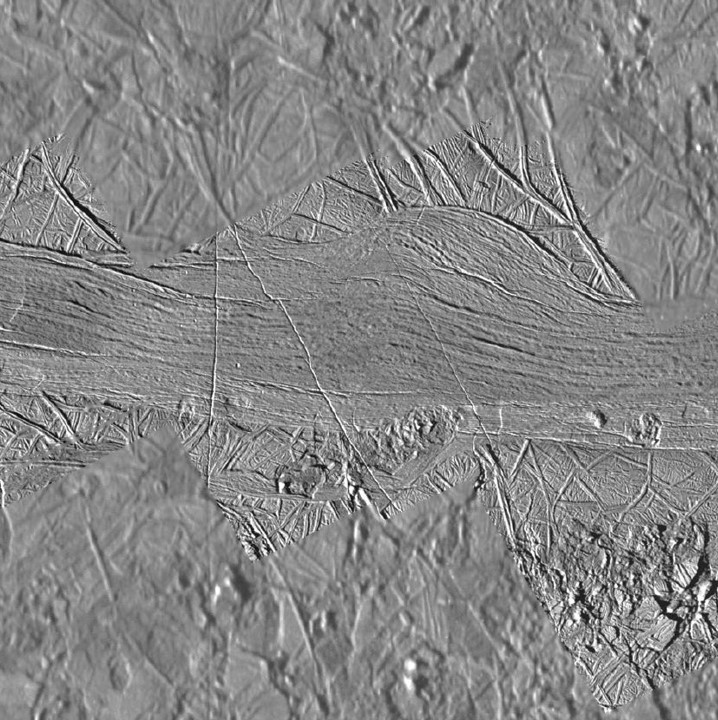}
		\label{fig:europa}
	}
}
\caption{Examples of low-feature terrains on extraterrestrial environments.}
\label{fig:illustration}
\end{figure*}

For example, consider the terrain at Purgatory Dune explored by Opportunity rover (see Figure \ref{fig:purgatory}). Visual feature detection for navigation is often insufficient to find adequate features in regolith. High-slip events and entrapment of the rovers are often experienced in terrains with unconsolidated sandy terrains~ \cite{arvidson2017mars, toupet2019terrain, wilcox1998sojourner, arvidson2014terrain}. Also, the majority of the \ac{VO} failures on Curiosity happened when the rover is at a sandy terrain with a few obvious unique features (66/94 by sol 2488)~\cite{rankin2020driving}.  Hence, the rover operation usually needs to be altered to detect unique terrain features, such as moving mast cameras~\cite{strader2020perception} to the point of interest or using wheel tracks left by the vehicle~\cite{Maimone07}. Degraded performance and unavailability of visual based systems are prone to increased localization drift on featureless terrains.
 
Apart from Martian exploration missions, the interest of the exploration of Europa has been increased~\cite{hand2018europa}. Although the knowledge of Europa’s surface is extremely limited, experiments for rover mobility purposes have been recently performed on salt-evaporites and icy terrains analogous to the Europa surface~\cite{reid2020autonomous}. Similar to the visually imperceptible unconsolidated sands buried under the thin cover of basaltic sands on Mars~\cite{arvidson2014terrain}, low-feature terrains on Europa are extremely challenging for vision-based systems and sparse features can degrade the perception performance leading to increased localization drift (see Figure \ref{fig:europa}). Therefore, given that slip detection is only performed by \ac{VO}-based methods in current rover operations\cite{toupet2019terrain, rankin2020driving}, the concern for providing continuous slip detection significantly increases when visual-based systems are unavailable.


\subsection{Contribution}

In our previous works, we employ two frameworks to leverage pseudo-measurement to provide reliable localization solution by using heuristically determined periodic stopping~\cite{kilic2019improved} and autonomous stopping~\cite{kilic2020}. The present work offers the contribution of detailing a proprioceptive slip detection technique that leverages the accurate velocity estimation from the localization framework detailed in our previous works. Current planetary rovers depend upon sufficiently detected and tracked features for exteroceptive slip detection. The proposed method does not depend on the visual characteristics of the environment and also only uses  the proprioceptive rover sensors already onboard. Therefore, it can be used as a complementary slip estimation technique when the visual sensor information is unavailable for the current and the upcoming planetary rover missions. The effectiveness of the proposed method is demonstrated with field tests in a perceptually degraded planetary-analog environment by qualitatively comparing with commercially off the shelf \ac{VIO} solution, wheel encoder based velocity estimation, and \ac{DGPS} velocity solutions.

\subsection{Outline of Paper}
The remainder of this paper is structured as follows: In Section \ref{sec:RW}, the literature on related works is reviewed. In Section \ref{sec:overview}, an overview of the methods used in the work is provided. The implementation of the methods for slip detection is detailed in Section \ref{sec:implementation}. Experimental setup, environment description and the discussion of the field experiments with the accuracy and efficiency of our method are provided in Section \ref{sec:field_experiments}. The conclusion of the work and the possible future research directions are summarized in Section \ref{sec:conclusion}.


\section{Related Work}

\label{sec:RW}

In general, wheel slippage can occur when the terrain traversed fails~\cite{iagnemma2004mobile} or when there is a kinematic incompatibility between wheels (i.e., different wheel speeds) encountered~\cite{gonzalez2018slippage}. Unexpected variances of terrains arise as non-systematic errors \cite{borenstein1996measurement}, and they can cause significant position errors. Also, driving across loose soil and sloped regions~\cite{c12} poses a substantial risk for wheel slippage. For example, if the rover traverses on a downward slope, the rover weight fraction in the movement's direction becomes greater, leading to an increased slippage. Vision-based approaches (e.g., VO) are mostly used to estimate the rover slip \cite{c12}, \cite{c64} \cite{Maimone07}. \ac{VO} is considered as an accurate and reliable source of information for slip estimation; however, it is computationally expensive for planetary rovers, especially when the rover is in motion. Using \ac{VO} substantially slows rover driving speed due to limited computational resources \cite{li2008characterization, toupet2019terrain}. Also, using \ac{VO} in a long period decreases the rover traversal rate since the rovers before Perseverance need to stop periodically to acquire images \cite{toupet2019terrain}. Even with the additional dedicated \ac{FPGA} processors~\cite{lentaris2015hw} and using the new enhanced Autonav for Perseverance rover~\cite{toupet2020rosbased}, the other limitations of \ac{VO} arise that it suffers from visually low-feature terrains and it relies on proper lighting conditions~\cite{strader2020perception}. Similarly, insufficiently detected and tracked features may lead to poor accuracy of motion estimate~\cite{c21}. This limitation can be a decisive factor when the terrain is covered in sand, making the terrain visually indistinguishable from sandy terrains. To alleviate some of the drawbacks of \ac{VO}, (e.g., poor illumination), there are some successful methods such as using vertically facedown cameras with LED lights \cite{nagatani2010development}, which can be considered for the future planetary robotic systems. However, providing a reliable localization estimation using the sensors already onboard without altering any rover operation is a challenging problem for current planetary rovers.
 
Martian soil is exceptionally challenging for traversability; even throughout a single drive, Mars rovers traverse on various terrains with different slopes~\cite{arvidson2014terrain}. Although the Martian terrain is not flat and contains a variety of local obstacle types \cite{arvidson2017mars}, the \ac{MSL} rover drives under the assumption of flat terrain if it does not run the Traction Control (TRCTL) algorithm, which is designed to reduce the rover wheel damage rate \cite{toupet2018traction, toupet2019terrain}. Various studies have modeled slip as a function of terrain geometry. A notable example of this is presented in ~\cite{angelova2006learning}, which uses a Mixture of Experts (MoE) structure to show the relationship between measured slip and visible terrain information. However, the wheel-terrain interactions are not always dictated by the apparent topsoil of the terrain~\cite{c21}. In particular, highly deformable sulfate-rich sands were concealed beneath the thin cover of basaltic sands on the Martian surface, which are not visually perceptible~\cite{arvidson2014terrain}. For example, \acp{MER} both became embedded into the soft surface of Mars~\cite{c18, c13} due to the significant amount of slip. In May 2009, Spirit became permanently entrapped in soft soil~\cite{c15}. Moreover, Curiosity faced a significant challenge to avoid sinking because of excessive wheel slippage on a sandy surface in Hidden Valley~\cite{c17, cunningham2017improving}. Due to greater compaction resistance, the rover wheels suffer sinkage-related slippage while traversing such terrains. A recent research has focused on exploiting the relation between slope and slip using proprioceptive sensors along with exteroceptive data~\cite{skonieczny2019data}. Although slope has a strong effect on slippage, wheel slippage may also be observed in relatively flat terrains if the rover wheels encounter kinematic incompatibility~\cite{gonzalez2018slippage}. This is often experienced when one of the rover's wheels crosses over a rock, it takes a longer path than the others, causing the other wheels to push forward~\cite{toupet2019terrain}. 

Reliable and continuous perception is the critical capability for resilient localization. Since no single observation can provide this, as previously mentioned, the vehicle can encounter critical navigation failures in perceptually degraded situations. However, proprioceptive sensing can be leveraged in addition to geometric and semantic information. A robotic system can utilize the information from some known conditions with proprioceptive sensing capabilities to provide a viable localization estimation in permanently shadowed regions, extremely bright areas, uniform, and visually indistinguishable terrains. Besides, the frequency of applying computationally expensive visual-based adjustments can be reduced by having a more consistent onboard proprioceptive localization. Making use of pseudo-measurements can be beneficial to improve the rover localization performance. There are several strategies to take advantage of pseudo-measurements including, but not limited to, zero velocity/angular rate~\cite{shin2005, grovebook, kilic2019improved, brossard2019rins}, zero displacement~\cite{grovebook, kilic2021zupt}, constant height/slope~\cite{klein2010pseudo}, and non-holonomicity \cite{niu2010using, grovebook, kilic2020}. In general, rovers are often in stationary mode~\cite{biesiadecki2006mars, Maimone07, rankin2020driving, strader2020perception}. Employing \ac{ZUPT} during stationary conditions is a well-known idea, which was initially popularized as a technique to assist inertial pedestrian navigation~\cite{foxlin2005, norrdine2016}. Publications that concentrate on pseudo-measurements more frequently adopt its effectiveness in standard road conditions~\cite{ramo, xiaofang2014, brossard2019rins}. Apart from utilizing pseudo-measurements to enhance localization for autonomous cars, using them can be considered in rough terrains as well. For example, given that planetary rovers stop more regularly than cars, \ac{ZUPT} is a well-suited application to deal with dead-reckoning drifts. Our previous studies have demonstrated that making use of pseudo-measurements can substantially improve the rover localization only with common sensors already onboard (i.e., an IMU and wheel encoders) without using any additional dedicated sensors or computational unit~\cite{kilic2019improved, kilic2020}. 


\section{Methods Overview } \label{sec:overview}

Apart from the exteroceptive slip detection in the current planetary rovers (e.g., Curiosity, Perseverance) that require camera outputs, this method utilizes the INS estimated velocity from our previous works~\cite{kilic2019improved, kilic2020}. Therefore, the accuracy of our proprioceptive slip detection relies on the reliability of INS state estimation. In this respect, this section provides the fundamental definitions and overviews of the methods to better grasp the underlying theoretical principles used in this paper.

\subsection{Proprioceptive Slip Detection} \label{sec:slip_detection}
The slip ratio is a simple indicator to observe the slip, which can be formulated in different ways. This formulation depends on the sensors used for estimating wheel velocity and body velocity~\cite{gonzalez2018slippage}. Observing the commanded distance and traversed distance, or commanded wheel velocity and actual velocity, can be given as examples. In this work, the slip ratio is obtained by comparing the difference between the velocity estimates from the filter and computed velocity based on the vehicle kinematics. 

The slip ratio~\cite{wong2008theory, amodeo2009wheel}, $s \in[-1,1]$, is defined as
\begin{equation}
\label{slip1}
s=\left\{\begin{array}{lll}
{1-\frac{v_x}{r \omega}} & { \text { (if }\omega \neq 0, v_x<r \omega,  \text { s $>$ 0 })} \\
{\frac{r \omega}{v_x}-1} & { \text { (if }v_x \neq 0, v_x>r \omega,  \text { s $<$ 0 })} \\
0 & { \text { (if } v_x = r\omega \lor v_x = \omega = 0  )}
\end{array}\right. 
\end{equation}
where ${v_x}$ is the translational velocity estimated from \ac{INS}, $r$ is the wheel radius, and $\omega$ is the wheel angular velocity estimated from the \ac{WO} measurements in our method. If slip ratio is equal to zero, it means the rover does not encounter any wheel slippage. Note that within this study, the wheel radius is assumed to remain constant, considering that the average wheel ground pressure is greater than the critical ground pressure and the wheel deformation is negligible \cite{siciliano2016springer}.

Slip ratio is usually distinguished as low, medium, and high slips. For example, upper and lower thresholds for these slip ratio classifications are given as low slip $(0 < |s| \leq 0.3)$, medium slip $(0.3 < |s| \leq 0.6)$, and high slip $(0.6 \leq |s|)$ in \cite{gonzalez2018slippage,gonzalez2019characterization, endo2021terrain}. It is reported that Curiosity rover traversal are forced to stop if a single significant slip measurement exceeds a threshold between 0.7 and 0.9 or if there is consecutive slip measurement between 0.4 and 0.7 based on the past experience with the terrain and testing \cite{arvidson2017mars}. Moreover, based on the terramechanic observations, 0.2 slip ratio is widely accepted as a significant slip threshold due to its effect on the drawbar pull. This effect can be observable as the rate of drawbar pull increasing with slip is higher between 0-0.2 slip range than 0.2 to 0.8 slip range\cite{skonieczny2019data}. This results in observing more sinkage in the 0.2 - 0.8 slip range. In this work, the slip ratio is classified in five classes as no slip $(s \approx 0)$, low slip $(0 < |s| \leq 0.2)$, medium slip $(0.2 < |s| \leq 0.4)$, high slip $(0.4 < |s| \leq 0.7)$, and extreme slip $(0.7 < |s| \leq 1.0)$ to include the slip ranges based on the previous observations and past experience from the Martian rovers.

Planetary rovers try to avoid significant slip; however, the slippage is almost inevitable on rough, sloped, and fine soil terrains. Experiencing a long term significant slip may cause catastrophic results, as in Spirit's mission; however, rovers can halt their driving before reaching that point and correct their route to reliably arrive the end goal. For this reason, the rover should be aware when experiencing a slip, even without using a visual-based system, given that most of the excessive slippage events happened in highly deformable and not visually perceptible unconsolidated sands buried under the thin cover of basaltic sands\cite{c21, arvidson2014terrain, c18, c13, c15, c17, cunningham2017improving, rankin2020driving}.

\subsection{Pseudo-Measurement Updates} \label{sec:pseudo_measurement}
Without external aiding, inertial sensor based state estimation inherently exhibits accumulated errors. A constant accelerometer bias causes a positioning error that grows quadratically in time, and a constant gyro rate bias results in a cubic error growth in position~\cite{siciliano2016springer}. Given that the wheel slippage mainly affects the wheel encoder based localization performance, a reliable inertial localization system can be used for detecting the velocity discrepancies between the rover (body) and wheel velocities. However, providing a reliable inertial localization system requires the calibration of the \ac{IMU} outputs. One way to calibrate the bias can be achieved by using the additional sensor outputs (e.g., magnetometers, sun sensors). However, magnetometers are usually not useful for obtaining global orientation when the planet of interest does not possess a global magnetic field. Similarly, sun sensors are only useful when the sun is within the sensor's field of view. Another way to calibrate the \ac{IMU} sensor biases can be achieved by utilizing pseudo-measurements in certain conditions from the sensors already onboard. For example, planetary rovers frequently stop for a variety of reasons, such as safety checks~\cite{rankin2020driving}, mast pointing~\cite{strader2020perception}, image processing \cite{Maimone07}, and conducting scientific experiments~\cite{biesiadecki2006mars}. Stopping is unavoidable even with the ideal case of the thinking-while-driving (TWD) approach, which has recently been developed to minimize how often the rover needs to stop for Perseverance~\cite{toupet2019terrain}. Since a rover is in stationary conditions in many instances, leveraging this state information is a natural fit for planetary robots.  

\begin{figure*}[htb]
\centerline
{
	\subfigure[]
	{
		\includegraphics[width=0.63\columnwidth]{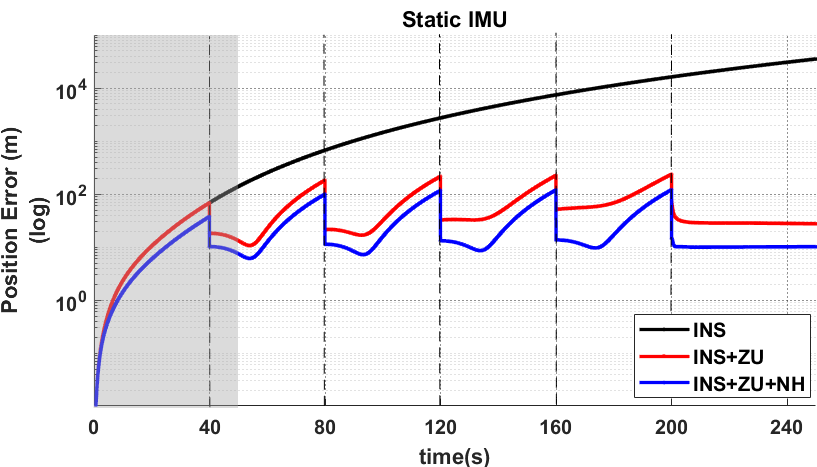}
		\label{fig:purgatory2}
	}
	\hspace{0.5\baselineskip}
	\subfigure[ ]
	{
		\includegraphics[width=0.31\columnwidth]{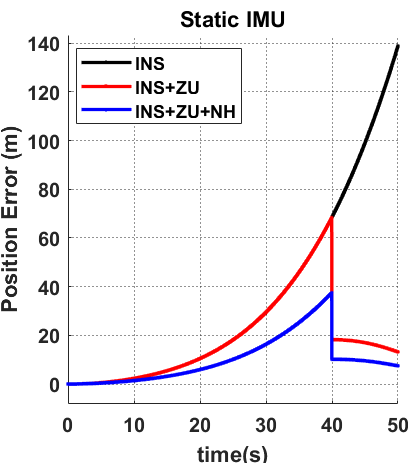}
		\label{fig:fig2}
	}
}
\caption{Toy example for using pseudo-measurements in the inertial positioning estimation. Left figure (a): INS based positioning with a static IMU under Earth gravity vector. Black line shows the INS only positioning estimation error, red line shows the error when using zero updates (labeled as ZU for both \ac{ZUPT} and \ac{ZARU}) in the INS estimation, the blue line shows the error when using zero updates and nonholonomic (labeled as NH) constraints in the INS estimation, and dashed black lines indicates the time step when the zero update is enabled for a step size based on the IMU data rate (0.02 s). The 3D positioning error is shown with a logarithmic scale. Right figure (b): The figure is the magnified part of the left figure (gray shaded area) between 0 to 50 seconds. The position error is plotted with a linear scale.  }
\label{fig:illustration2}
\end{figure*}

Pseudo-measurements are the constraints that available to use in certain kinematic and physical conditions. These pseudo-measurements can be applied as a measurement update to enforce constraints on the states of a system and can be used in the system state estimation process in a cost-effective way because the information is mostly gathered from the sensors already on-board. A toy example to demonstrate the pseudo-measurement capability of reducing error growth in the INS-based localization is given in Fig.~\ref{fig:illustration2}. In this figure, a static IMU output with 50~Hz data rate under the effect of Earth's gravity field is processed with and without pseudo-measurement updates. Since the used IMU is stationary for this toy example, we were able to control when to enable pseudo-measurement updates in the state estimation. For example, \ac{ZUPT} and \ac{ZARU} (zero updates) are enabled for one step size of the estimation (0.02~s), every 40 seconds. Non-holonomic constraint updates are used in each time step of the INS estimation. After 200th second, zero updates are kept active. While it may be intuitive that these constraints provide information useful for calibrating \ac{IMU} sensor biases, this figure also illustrates that the position errors are further reduced to some extent after an update. This is due to the fact that correlation of IMU biases to position errors are modeled and integrated over time in the INS process noise covariance matrix. In the following sections, implementation details of the pseudo-measurement updates in an error state \ac{EKF} for INS based state estimation are given.     

\paragraph{Zero Velocity Update}
\ac{ZUPT} can bound the velocity error, calibrate IMU sensor biases, and limit the rate of INS localization error growth \cite{grovebook}. In a planetary mission, the localization system needs to be computationally tractable. Leveraging \ac{ZUPT} in a localization framework does not require any dedicated sensor or complicated processes except the acquisition of the \ac{IMU} and wheel encoder data. Therefore, this can provide computationally efficient and accurate real-time rover localization capability with the operations which are already available most of the time for rovers. This is a particularly desirable consequence of using ZUPT in planetary robotics, as computational resources in planetary rovers are limited by radiation-hardened hardware. IMU sensor outputs are governed by sensor errors during the zero velocity. The measurement noise covariance describes the variance and covariance of the nominally-zero velocity due to vibration and disturbances. This fact is used when performing zero velocity update and the measurements fed into error state \ac{EKF} to reduce the localization error growth of the system. Because the error state model maintains the correlation between the position and velocity errors of the error covariance matrix, this procedure helps in decreasing the INS positioning error growth from cubic to linear \cite{mather2006man}. This allows ZUPT to restrict error increase, assist in the determination of biases which can be used to decrease future error growth, and offset the majority of position drift since the last navigation stop. Improving localization performance by using ZUPT is detailed in our previous works \cite{kilic2019improved, kilic2020}.

\paragraph{Zero Angular Rate Update}
During stationary conditions, \ac{ZUPT} can bound the roll and pitch errors; however, heading error may accumulate rapidly due to poor observation of heading~\cite{shin2005,wahlstrom2020fifteen}. In this case, a \ac{ZARU} can be performed to decrease the heading drift during a \ac{ZUPT}. Similar to \ac{ZUPT}, the idea is to use pseudo-measurement updates in the error state filter with angular rate error~\cite{jimenez2010indoor}. In order to use a \ac{ZARU}, the angular rate of the rover should be zero. This can be identified by comparing the standard deviation of the recent yaw-rate gyro measurements, steering-angle commands (considering steerable wheels), and the yaw rate obtained from wheel odometry.

\paragraph{Non-Holonomic Motion Constraints}
The non-holonomic motion constraints can be used when the rover wheels do not move vertical or lateral to the traversal surface. This can be interpreted as that these constraints become invalid when the rover experiences a slip on its sideways or a wheel loses its contact to the surface for an extended period. For skid-steer rovers, even the rover wheels are not steerable, these constraints are broken when a skid-steer robot experience significant lateral slip or during turning motion. However, for steerable wheels, as in double Ackermann or four-wheel-steering modes, these constraints can only be used during driving with zero velocity along the rotation axis of any of rover wheels~\cite{diss}.

\section{Implementation} \label{sec:implementation}

\begin{figure}[b!]
    \centering
    \includegraphics[width=0.8\linewidth]{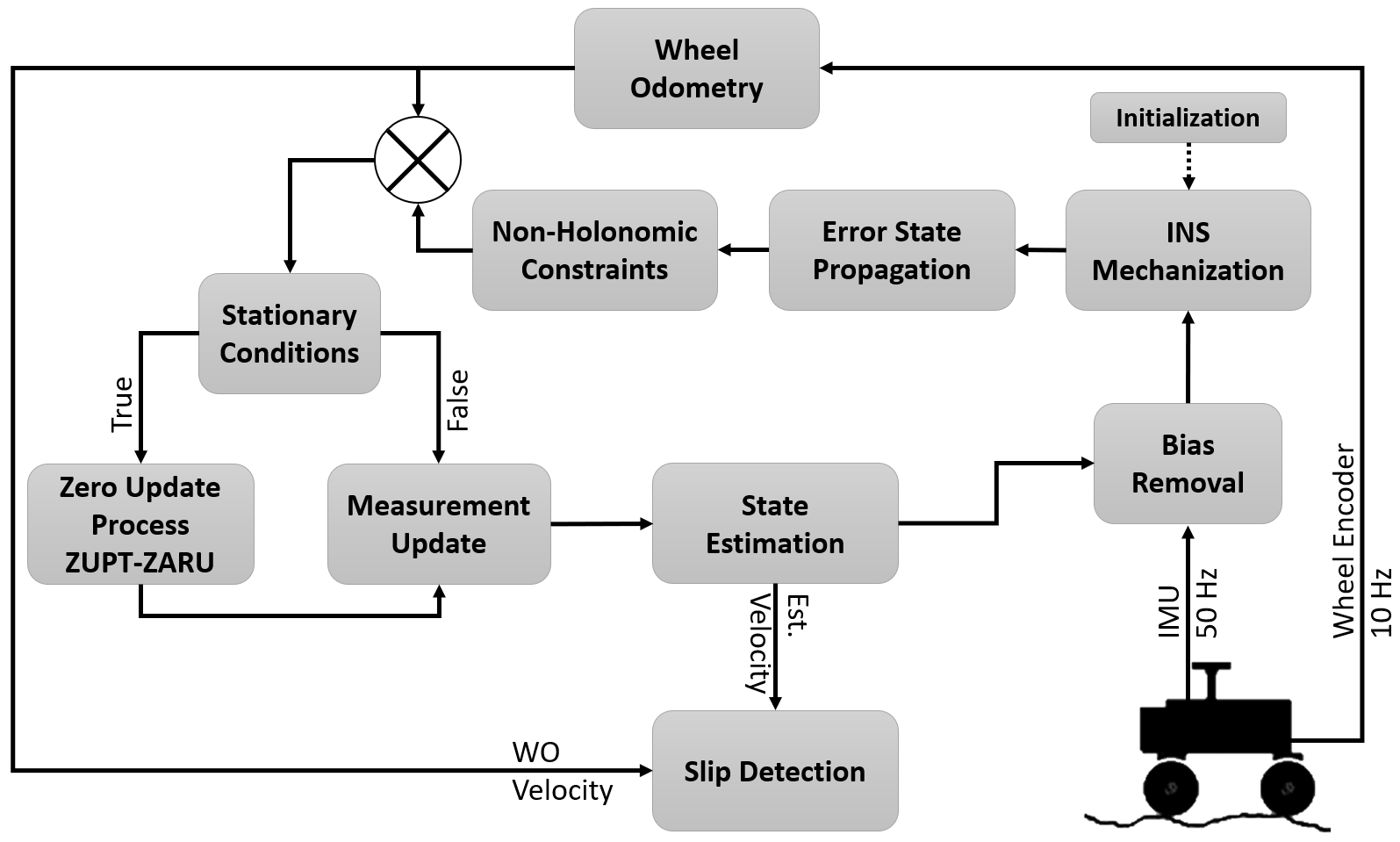}
    \caption{A framework depiction of the proposed method. After initialization, the IMU observations are utilized in the mechanization process and the propagation of the error state. Pseudo-measurements (ZUPT, ZARU, and Non-Holonomic constraints) can be introduced to the state estimation process in certain kinematic and physical conditions that detailed previously. Slip detection is performed by comparing the difference between the velocity estimates from the filter and computed velocity based on the vehicle kinematics.  }
    \label{fig:framework}
\end{figure}

 In this section, we present further details for our implementation to allow the reader to more easily replicate our method. The implementation process includes a comprehensive formulation of the \ac{INS} mechanization, state estimation through an error-state \ac{EKF}, using pseudo-measurements in the error state, and slip detection mechanism.  A depiction of the framework is given in Fig.~\ref{fig:framework}.
 
 \subsection{INS Mechanization}
  The navigation equations are implemented in a locally level navigation frame following the strapdown INS mechanization process as detailed in  \cite{savage1998strapdown2,savage1998strapdown1,titterton2004strapdown} with closely following the formulation and notation provided in~\cite{grovebook}, and used in~\cite{kilic2019improved,kilic2020}.  

\paragraph{Attitude Update:}The attitude estimation in a locally level navigation frame implementation can be expressed as the body to navigation frame coordinate transformation matrix. The attitude update is given as
\begin{equation}
\label{eq:su2}
\begin{aligned}
{\mathbf{C}_{b}^{n}}^{\scriptscriptstyle (+)}\approx{} & {\mathbf{{C}}_{b}^{n}}^{\scriptscriptstyle(-)}  \bigl (\mathbf{I} +[ \boldsymbol{\omega}_{ib}^{b} \times] \tau_{s} \bigr ) -\bigl ({[\boldsymbol{\omega}_{ie}^{n}}^{\scriptscriptstyle(-)}\times] +{[\boldsymbol{\omega}_{en}^{n}}^{\scriptscriptstyle(-)}  \times]  \bigr ){\mathbf{C}_{b}^{n}}^{\scriptscriptstyle(-)} \tau_{s}
\end{aligned}
\end{equation}
where  $\mathbf{C}_{b}^{n} \in \mathbb{R}^{3 \times 3}$ is the coordinate transformation matrix from the body frame to the locally level frame, $\mathbf{I}$ is identity matrix, $\mathbf{\omega}_{ib}^{b}$ is the \ac{IMU} angular rate measurement, $\mathbf{\omega}_{ie}^{n}$ is the planet's rotation vector represented in the locally level frame, $\mathbf{\omega}_{en}^{n}$ is the transport term, $\tau_{s}$ is the \ac{IMU} sampling interval, and notation ``$\times$'' stands for the skew-symmetric matrix of the vector.

\paragraph{Velocity Update:}Assuming that the variations of the acceleration due to gravity, Coriolis, and transport rate terms are all negligible over the integration interval, the velocity update is given as,
\begin{equation}
\begin{aligned}
{\mathbf{v}_{eb}^{n}}^{\scriptscriptstyle(+)} \approx {} & {\mathbf{v}_{eb}^{n}}^{\scriptscriptstyle(-)} +\bigl ( \mathbf{f}_{ib}^{n}+\mathbf{g}_{b}^{n}(L_b^{\scriptscriptstyle(-)} ,h_b^{\scriptscriptstyle(-)} ) - ( [{\boldsymbol{\omega}_{en}^{n}}^{\scriptscriptstyle(-)} \times]  +2{[\boldsymbol{\omega}_{ie}^{n}}^{\scriptscriptstyle(-)} \times]){\mathbf{v}_{eb}^{n}}^{\scriptscriptstyle(-)} \bigr )\tau_{s} 
\end{aligned}
\end{equation}
where $\mathbf{v}_{eb}^{n} \in \mathbb{R}^3 $ is the velocity update, $\mathbf{f}_{ib}^{n} \in \mathbb{R}^3$ is the specific force measurements from the \ac{IMU} acceleration sensors, $\mathbf{g}_{b}^{n}$ is the gravity vector. The velocity estimation is given as planet referenced in locally level navigation frame.

\paragraph{Position Update:}Assuming the velocity variation is linear over the integration interval, the position update in the curvilinear form (latitude, longitude, height) is given as
\begin{equation}
h_b^{(+)} =h_b^{\scriptscriptstyle(-)} - \bigl({{\mathbf{v}_{eb,D}^{n}}^{\scriptscriptstyle(-)}  +{\mathbf{v}_{eb,D}^{n}}^{\scriptscriptstyle(+)}}  \bigr) \frac{\tau_{s}}{2}
\end{equation} 

\begin{equation}
\begin{aligned}
L_b^{\scriptscriptstyle(+)} =L_b^{\scriptscriptstyle(-)} +\frac{\tau_{s}}{2} \frac{{\mathbf{v}_{eb,N}^{n}}^{\scriptscriptstyle(-)} }{R_N(L_b^{(-)} )+h_b^{\scriptscriptstyle(-)} }+\frac{\tau_{s}}{2}\frac{{\mathbf{v}_{eb,N}^{n}}^{\scriptscriptstyle(+)} }{R_N(L_b^{\scriptscriptstyle(-)} )+h_b^{\scriptscriptstyle(+)} } 
\end{aligned}
\end{equation}

\begin{equation}
\begin{aligned}
\lambda_b^{\scriptscriptstyle(+)} =\lambda_b^{\scriptscriptstyle(-)} +\frac{\tau_{s}}{2} \frac{{\mathbf{v}_{eb,E}^{n}}^{\scriptscriptstyle(-)} }{\bigl(R_E(L_b^{\scriptscriptstyle(-)} )+h_b^{\scriptscriptstyle(-)} \bigr)cosL_b^{\scriptscriptstyle(-)} } +\frac{\tau_{s}}{2}\frac{{\mathbf{v}_{eb,E}^{n}}^{\scriptscriptstyle(+)}}{\bigl(R_P(L_b^{(-)})+h_b^{\scriptscriptstyle(+)}\bigr)cosL_b^{\scriptscriptstyle(+)}}
\end{aligned}
\end{equation}
where $h_b$, $L_b$ and $\lambda_b$ are updated position estimates (expressed in terms of height, latitude, and longitude, respectively), $R_N$ is the variation of the meridian, and $R_P$ is transverse radii of curvature.
 \subsection{Error State Extended Kalman Filter} \label{sec:esekf}
 The Kalman filter is a linear estimator and cannot be used directly to estimate states that are non-linear functions of either the measurements or the control inputs. The key concept in the extended Kalman filter is the idea of linearization a non-linear system about the current best estimate \cite{jazwinski2007stochastic}. 
In the error state \ac{EKF}, the error state is estimated instead of the total state. Then the estimate of this error state can be used to correct the total state~\cite{roumeliotis1999circumventing, grovebook}. The total state vector is given as
\begin{equation}
\mathbf{x}^{n}={\biggl(
\mathbf{\Psi}_{nb}^{n} \ \ 
\mathbf{v}_{eb}^{n} \ \ 
\mathbf{p}_{b} 
\biggr )}^{\text{T}} 
\end{equation}

\paragraph{Error State Model}
The error state, $\delta \mathbf{x}^{n} \in \mathbb{R}^{15}$, is constructed in a local navigation frame, 
\begin{equation}
\label{errorstate}
\delta \mathbf{x}^{n} = {\biggl(
\delta\mathbf{ \Psi}_{nb}^{n} \ \ 
\mathbf{\delta v}_{eb}^{n} \ \ 
\delta\mathbf{p}_{b} \ \ 
\mathbf{b}_a \ \ 
\mathbf{b}_g
\biggr )}^{\text{T}}, 
\quad 
\delta \mathbf{p}_{b}={\biggl(
\delta {L}_{b} \ \ 
\delta {\lambda}_{b} \ \
\delta {h}_{b}
\biggr)}^{\text{T}}
\end{equation}
where, $\delta\mathbf{ \Psi}_{nb}^{n} \in \mathbb{R}^3$ is the attitude error, $\mathbf{\delta v}_{eb}^{n} \in \mathbb{R}^3$ is the velocity error, $\delta\mathbf{p}_{b}\in \mathbb{R}^3$ is the position error, $\mathbf{b}_a \in \mathbb{R}^3$ is the \ac{IMU} acceleration bias, and $\mathbf{b}_g \in \mathbb{R}^3$ is the \ac{IMU} gyroscope bias.
The position error is expressed in terms of the latitude, longitude, and height, respectively.
\paragraph{INS System Matrix}
The Jacobian of the error-state equations is used to compute the \ac{INS} system matrix.   The system matrix and the \ac{STM} are built once the time derivatives of the error-state equations are defined. The errors are then converted into the local navigation frame, and the velocity error's time derivative is calculated by including the transport rate component.

The system matrix can be given as; 
\begin{equation}\label{eq:F}
\mathbf{F}^{n}=\left[\begin{array}{ccccc}
\mathbf{F}_{11}^{n} & \mathbf{F}_{12}^{n} & \mathbf{F}_{13}^{n} & \mathbf{0}_{3} & \hat{\mathbf{C}}_{b}^{n} \\
\mathbf{F}_{21}^{n} & \mathbf{F}_{22}^{n} & \mathbf{F}_{23}^{n} & \hat{\mathbf{C}}_{b}^{n} & \mathbf{0}_{3} \\
\mathbf{0}_{3} & \mathbf{F}_{32}^{n} & \mathbf{F}_{33}^{n} & \mathbf{0}_{3} & \mathbf{0}_{3} \\
\mathbf{0}_{3} & \mathbf{0}_{3} & \mathbf{0}_{3} & \mathbf{0}_{3} & \mathbf{0}_{3} \\
\mathbf{0}_{3} & \mathbf{0}_{3} & \mathbf{0}_{3} & \mathbf{0}_{3} & \mathbf{0}_{3}
\end{array}\right].
\end{equation}
The elements of the \ac{INS} system matrix are provided from Equation~\ref{F11} to \ref{F33}. Using the time derivatives of the error-state equations, the state transition model can be assumed as;
\begin{equation}
    \boldsymbol{\Phi}_k \approx e^{\mathbf{F}_k \tau_s} = \sum_{\alpha=0}^{\infty} \frac{\mathbf{F}_{k}^{\alpha} \tau_{s}^{i}}{\alpha !}.
\end{equation}
Neglecting the higher order terms after the first order, the elements of the STM for discrete time can be approximated to
\begin{equation}\label{phi}
\boldsymbol{\Phi}_{k}^{n} \approx\left[\begin{array}{ccccc}
\mathbf{I}_{3}+\mathbf{F}_{11}^{n} \tau_{s} & \mathbf{~F}_{12}^{n} \tau_{s} & \mathbf{~F}_{13}^{n} \tau_{s} & \textbf{0}_{3} & \hat{\mathbf{C}}_{b}^{n} \tau_{s} \\
\mathbf{~F}_{21}^{n} \tau_{s} & \mathbf{I}_{3}+\mathbf{F}_{22}^{n} \tau_{s} & \mathbf{~F}_{23}^{n} \tau_{s} & \hat{\mathbf{C}}_{b}^{n} \tau_{s} & \textbf{0}_{3} \\
\textbf{0}_{3} & \mathbf{~F}_{32}^{n} \tau_{s} & \mathbf{I}_{3}+\mathbf{F}_{33}^{n} \tau_{s} & \textbf{0}_{3} & \textbf{0}_{3} \\
\textbf{0}_{3} & \textbf{0}_{3} & \textbf{0}_{3} & \mathbf{I}_{3} & \textbf{0}_{3} \\
\textbf{0}_{3} & \textbf{0}_{3} & \textbf{0}_{3} & \textbf{0}_{3} & \mathbf{I}_{3}
\end{array}\right]
\end{equation}

\paragraph{Propagation/Prediction}
In the propagation step, the error state given in Equation~(\ref{errorstate}) is propagated using the \ac{STM}, such that
\begin{equation}
\delta \check{\mathbf{x}}_{k} = \boldsymbol{\Phi}_{k-1}\delta \hat{\mathbf{x}}_{k-1}  
\end{equation}

Similarly, the error covariance matrix $\mathbf{P} \in \mathbb{R}^{15 \times 15}$ is propagated through if pseudo-measurements are unavailable
\begin{align}
\check{\mathbf{P}}_{k}&=\boldsymbol{\Phi}_{k-1} \mathbf{P}_{k-1} \boldsymbol{\Phi}_{k-1}^{\text{T}}+\mathbf{L}_{k-1} \mathbf{Q}_{k-1} \mathbf{L}_{k-1}^{\text{T}}
\end{align}
where $\mathbf{Q}_{k}$ is the process noise covariance, $\mathbf{L}_{k}$ is the process noise related Jacobian, and it is identity in our case since the noise is assumed to be additive.

The process noise covariance matrix can be defined by the random walk of the velocity error due to noise on the accelerometer specific-force measurements and random walk of the attitude error due to noise on the gyro angular rate measurements \cite{grovebook}. Integrating the power spectral densities of the accelerometer and gyroscope noise over the state propagation interval, the INS system noise covariance matrix can be given with closely following the notation in~\cite{grovebook};

\begin{equation}
\mathbf{Q}_{k}^{n}=\left[\begin{array}{ccccc}
\mathbf{Q}_{11}^n & \mathbf{Q}_{21}^{n \text{T}} & \mathbf{Q}_{31}^{n \text{T}} & \textbf{0}_{3} & \frac{1}{2} S_{b g d} \tau_{s}^{2} \hat{\mathbf{C}}_{b}^{n} \\
\mathbf{Q}_{21}^{n} & \mathbf{Q}_{22}^{n} & \mathbf{Q}_{32}^{n \text{T}} & \frac{1}{2} S_{b a d} \tau_{s}^{2} \hat{\mathbf{C}}_{b}^{n} & \frac{1}{3} S_{b g d} \tau_{s}^{3} \mathbf{F}_{21}^{n} \hat{\mathbf{C}}_{b}^{n} \\
\mathbf{Q}_{31}^{n} & \mathbf{Q}_{32}^{n} &\mathbf{Q}_{33}^{n} & \mathbf{Q}_{34}^{n} & \mathbf{Q}_{35}^{n} \\
\textbf{0}_{3} & \frac{1}{2} S_{b a d} \tau_{s}^{2} \hat{\mathbf{C}}_{n}^{b} & \mathbf{Q}_{34}^{n \text{T}} & S_{b a d} \tau_{s} \mathbf{I}_{3} & \textbf{0}_{3} \\
\frac{1}{2} S_{b g d} \tau_{s}^{2} \hat{\mathbf{C}}_{n}^{b} & \frac{1}{3} S_{b g d} \tau_{s}^{3} \mathbf{F}_{21}^{n} \hat{\mathbf{C}}_{n}^{b} & \mathbf{Q}_{35}^{n \text{T}} & \textbf{0}_{3} & S_{b g d} \tau_{s} \mathbf{I}_{3}
\end{array}\right]\label{eq:qmatrix}\end{equation}
\hfill \break
where $S_{rg}$, $S_{ra}$, $S_{bad}$, and $S_{bgd}$ are the power spectral densities of the gyro random noise, accelerometer random noise, accelerometer bias variation, and gyro bias variation, respectively \cite{grovebook}. The elements of the INS system noise covariance matrix in Equation~(\ref{eq:qmatrix}) are given in Equations~\ref{Qvalues} to \ref{Trp}.

The optimal gain can be calculated as;
\begin{equation}
\mathbf{K}_{k}=\check{\mathbf{P}}_{k} \mathbf{H}_{k}^{T}\left(\mathbf{H}_{k} \check{\mathbf{P}}_{k} \mathbf{H}_{k}^{T}+\mathbf{R}_{k}\right)^{-1}\label{eq:kalmangain}
\end{equation}
where $\mathbf{H}_{k}$ is the measurement matrix and $\mathbf{R}_{k}$ is the measurement noise covariance. The estimate of the error state can be given as; 
\begin{equation}
\delta \hat{\mathbf{x}}_{k}=\mathbf{K}_{k}\left(\mathbf{\delta z}_{k}-\mathbf{H}_{k} \delta \check{\mathbf{x}}_{k}\right)
\end{equation}
where $\mathbf{z}_{k}$ is the corresponding available pseudo-measurement innovation that is given in the following for each type of updates used in this work.
\paragraph{Pseudo-Measurement Updates}

\paragraph{\textit{Zero Velocity Update:}}  The measurement innovation for \ac{ZUPT} is
\begin{equation}
\mathbf{\delta z}_{ZV,k}^{\gamma -}=-\mathbf{\hat{v}}_{eb,k}^{\gamma} \quad \gamma  \in e,n
\end{equation}
and the measurement matrix is
\begin{equation}
\mathbf{H}_{ZV,k}^{\gamma}=\biggl( \mathbf{0}_3 \ \ {\scriptstyle-}\mathbf{I}_3 \ \ \mathbf{0}_3 \ \ \mathbf{0}_3 \ \ \mathbf{0}_3 \biggr) \quad \gamma \in e,n .
\end{equation}
Although \ac{ZUPT} does not provide absolute position information, the Kalman filter system model builds up information on the correlation between the velocity and position errors in the off-diagonal elements of the error covariance matrix, $\mathbf{P}$, and the cubic error growth for positioning is reduced to linear. This enables a \ac{ZUPT} to correct most of the position drift since the last measurement update (see Equations~(\ref{eq:F}) and (\ref{phi})). 

\paragraph{\textit{Zero Angular Rate Update:}}
The measurement innovation for a \ac{ZARU} can be given independent of the coordinate frames used for position, velocity and attitude states as
\begin{equation}
\mathbf{\delta z}_{ZA,k}^{-}=-\mathbf{\hat{\omega}}_{ib,k}^{b} \quad 
\end{equation}
and the measurement matrix is
\begin{equation}
\mathbf{H}_{ZA,k}=\biggl( \mathbf{0}_3 \ \ \mathbf{0}_3 \ \ \mathbf{0}_3 \ \ \mathbf{0}_3 \ \ {\scriptstyle-}\mathbf{I}_3  \biggr). 
\end{equation}

\paragraph{\textit{Non-Holonomic Motion Constraints:}}

The rover velocity constraints can be applied as a pseudo-measurement update. Following the notation in~\cite{grovebook}, the measurement update for the non-holonomic motion constraints can be expressed as
\begin{equation}
\mathbf{\delta z}_{RC}^{n}=-\begin{pmatrix}
0 & 1 & 0\\ 
0 & 0 & 1
\end{pmatrix} (\mathbf{C}_{n}^{b} \mathbf{v}_{eb}^{n} -[\mathbf{\omega}_{ib}^{b} \times] |\mathbf{L}_{wb}^{b}|) 
\end{equation}
where $\mathbf{L}_{wb}^{b} \in \mathbb{R}^3 $ is the lever arm from the non-steerable wheel frame to the body frame and $\mathbf{\omega}_{ib}^{b} \in \mathbb{R}^3$ is angular rate measurement. In our setup, both rear and front wheels are non-steerable (skid-steer), and the IMU has the same distance between rear and front wheels. For steerable vehicles, this cannot be generalized as zero lateral velocity and the lever arm should be taken properly to use non-holonomic motion constraints. Corresponding measurement matrix may be approximated as

\begin{equation}
\mathbf{H}_{RC}^{n}=\begin{pmatrix}
\mathbf{0}_{2,3} & \begin{bmatrix}-\mathbf{H}_{l} \\ -\mathbf{H}_{v}\end{bmatrix} &\mathbf{0}_{2,3}  &\mathbf{0}_{2,3} &\mathbf{0}_{2,3}
\end{pmatrix}
\end{equation}
where $\mathbf{H}_{l}$ is lateral constraint part. and $\mathbf{H}_{v}$ is the vertical part of the measurement matrix. 
\begin{equation}
\begin{bmatrix}-\mathbf{H}_{l} \\ -\mathbf{H}_{v}\end{bmatrix} = \begin{pmatrix}
0 & 1 & 0 \\ 0 & 0 & 1 \end{pmatrix} \mathbf{C}_{n}^{b}.
\end{equation}

\paragraph{State Correction}
After having the error state estimation, it can be used to correct the state. Starting from coordinate transformation matrix correction,
\begin{equation}
    {\hat{\mathbf{C}}_{n}^{b}}=(\mathbf{I}_3 -[\delta{\mathbf{\Psi}}_{k}) \times]) {\mathbf{\check{C}}_{b}^{n \text{T}}}.
\end{equation}
Then, the attitude correction can be given as;
\begin{equation}
\hat{\boldsymbol{\Psi}}_{n b}^{n}=\left[\begin{array}{c}
\operatorname{atan} 2\left(\hat{\mathbf{C}}_{n(3,2)}^{b}, \hat{\mathbf{C}}_{n(3,3)}^{b}\right) \\
\operatorname{asin}\left(-\hat{\mathbf{C}}_{n(3,1)}^{b}\right) \\
\operatorname{atan} 2\left(\hat{\mathbf{C}}_{n(2,1)}^{b}, \hat{\mathbf{C}}_{n(1,1)}^{b}\right)
\end{array}\right]
\end{equation}
The velocity and position corrections are given as;
\begin{align}
    \hat{\mathbf{v}}_{eb}^{n}&=\check{\mathbf{v}}_{eb}^{n}+\delta{\mathbf{v}}_{eb}^{n}\\
    \hat{\mathbf{p}}_{b}&=\check{\mathbf{p}}_{b}+\delta{\mathbf{p}}_{b}.
\end{align}

Finally, the error covariance matrix can be corrected as;
\begin{equation}
\hat{\mathbf{P}}_{k}=\left(\mathbf{I}-\mathbf{K}_{k} \mathbf{H}_{k}\right) \check{\mathbf{P}}_{k}.
\end{equation}

\subsection{Slip Detection} \label{sec:slip_detection}
 Considering the Equation~\ref{slip1} the proprioceptive slip detection accuracy depends on the quality of the velocity estimation from the filter. In an extreme case with a perfect INS state estimation (i.e., no INS drift), rover localization would not be affected of this slip. In fact, wheel slippage only affects the wheel odometry estimation considering the problem in the localization perspective. However, an INS-based dead-reckoning system is prone to drift in the real world scenarios. Using pseudo-measurement updates can significantly improve the \ac{INS} estimated velocity accuracy. Consequently, improved \ac{INS} velocity estimation can be used to detect slippage by comparing it with wheel encoder based wheel velocities when \ac{VO} is not available. The translational velocity is obtained with transforming the \ac{INS} velocity from navigation frame to body frame such that
\begin{equation}
[v_x,v_y,v_z]^T=C_n^b {\mathbf{v}_{eb}^{n}}.
\end{equation}


\section{Field Experiments} \label{sec:field_experiments}

\subsection{Experimental Setup} \label{sec:experimental_setup}

\paragraph{Robot Description}
WVU Pathfinder is a lightweight, four-wheeled, skid-steered testing platform. The rover drive-train is a double-bogie (as opposed to a rocker-bogie) system, where the two wheels on one side are attached to the same bogie, and the robot body and other bogie are constrained by a rotational differential bar. Left and right wheel pairs are controlled by motor controllers and mounted on each side of the rover. It has $24~cm$ diameter polyurethane wheels that each wheel is capable of carrying maximum of 40 kg payload. Even these wheels are made of deformable polyurethane material, they are assumed as rigid wheels since their inflation pressure are high (4 PSI) and the testing platform is lightweight \cite{siciliano2016springer}. Pathfinder track-width is $0.685~m$, and wheelbase is $0.544~m$. The rover is $15~kg$, and its maximum speed is $0.8~m/s$. 
\paragraph{Robot Sensors}
Different proprioceptive and exteroceptive sensor modalities such as \ac{IMU}, wheel encoders, and tracking cameras, as well as GNSS receivers and antennas, are leveraged to evaluate the results with reference solutions and for comparisons. To measure the acceleration and rate gyro signals, 16495~(ADIS 16495-2BMLZ) 6-degree of freedom (DOF) \ac{IMU} model, includes a tri-axis gyroscope and a tri-axis accelerometers, is utilized. ADIS 16495-2BMLZ IMU in-run bias and angular random walk values are 1.6 deg$/$hr, 0.1 deg$/\sqrt{\text{hr}}$ for the gyroscope, and 3.2~$\mu$g, 0.008~m$/$sec$/\sqrt{\text{hr}}$ for the accelerometer, respectively. Wheel odometry estimations are used as a standalone dead-reckoning solution for comparisons and as an aiding measurement update to the inertial localization system. The \ac{WO} inputs are generated by quadrature Hall effect encoder feedback with 47,000~pulses$/$m for Pathfinder. Global positioning information is collected with dual-frequency Novatel OEM-615 GNSS receiver and L1/L2 Pinwheel antenna that mounted to the rovers. The GNSS solution is used for the initialization of the inertial localization framework. The post-processed high precision solution (e.g., DGPS) is leveraged for comparisons and truth generation for this work. Additionally, a commercially-of-the-shelf system, Intel RealSense T265 tracking camera is used for localization performance comparisons. This system includes two fisheye lens sensors, an IMU, and an Intel visual processing unit (VPU). 

\begin{figure}[htb]
    \centering
    \includegraphics[width=0.64\linewidth]{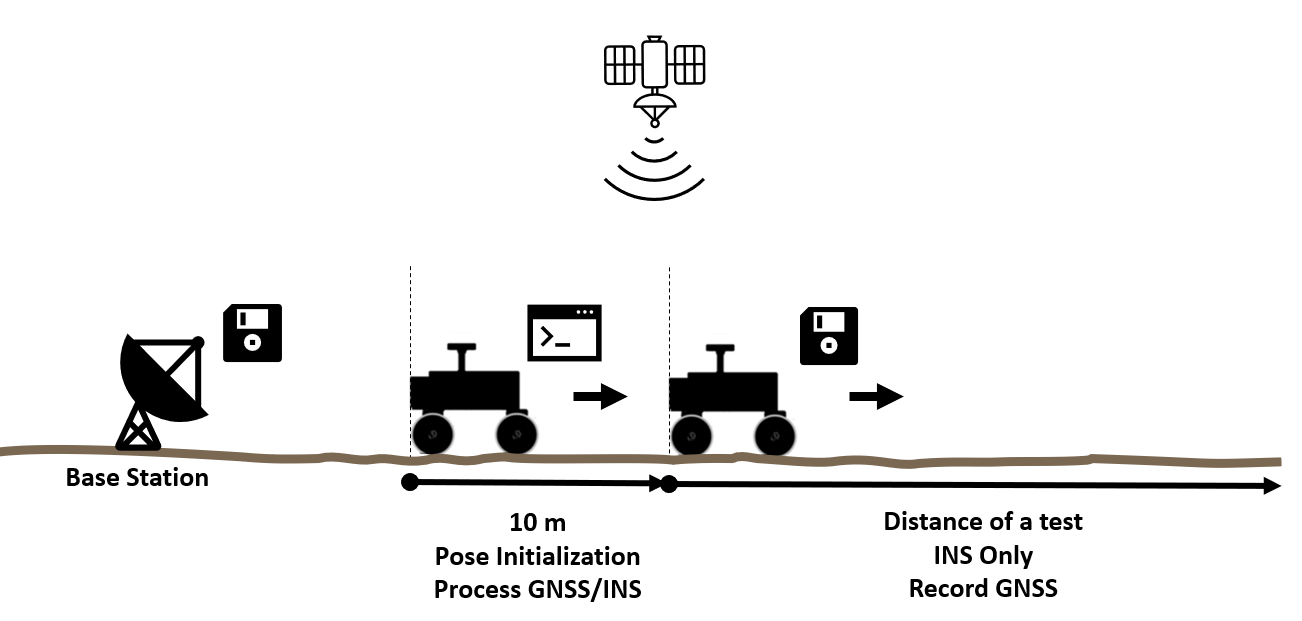}
    \caption{Demonstration of the initial pose generation.  }
    \label{fig:initialization}
\end{figure}

An IntelCore i7-8650U CPU used in the rover to run the developed software and collect the processed position data from the tracking system.  \ac{ROS} framework is used for the software development, data processing, and data collection. One lesson learned from this experimental work was the need for careful design of sensor update rates in consideration of real-time processing with \ac{ROS}. For example, our IMU was capable of providing 200 Hz measurements; however, processing at this speed rate led to an increased potential of missing sensor callbacks. Due to this reason, in this work, we decided to reduce the IMU processing rate from 200 Hz to 50 Hz and it was observed that using 50 Hz IMU processing rate did not lead missing sensor messages in \ac{ROS}. The proposed method, tracking system, wheel odometry, and \ac{GPS} information are stored in the computer and post-processed for comparisons and the datasets are made publicly available\footnote{https://dx.doi.org/10.21227/vz7z-jc84}.
\paragraph{Truth Reference}
 Integer-ambiguity-fixed carrier-phase \ac{DGPS} is used to determine a truth reference solution. The  setup for the DGPS solution consisted of two dual-frequency GNSS receivers and dual-frequency antennas, with one set mounted on a static base station and another affixed on top of the test rover platforms. Both receivers recorded 10~Hz carrier-phase and GPS pseudorange readings during the tests. These external GNSS measurements are utilized to generate the \ac{DGPS} solution during post-processing using RTKLIB 2.4.2~\cite{rtklib} software library. The \ac{DGPS} solution, which provides a cm-to-dm expected level of accuracy~\cite{gps}, is adopted as the truth in comparison analyses.    
\paragraph{Initial Pose Condition Estimation}
 A loosely-coupled GNSS-INS sensor fusion approach is used to initialize the pose prior to tests. The procedure starts with driving straight for a short distance ($\sim 10m$) to assess the absolute heading estimate. Then, the rover remains stationary for a reasonable time ($\sim30s$) to initialize position before starting experiments.  The rover is assumed as it starts the driving on a flat region (i.e., initial roll and pitch are equal to zero degrees). This assumption is only applied for the initialization process. A depiction of this process is given in Fig.~\ref{fig:initialization}. 

\subsection{Environment Description}
\begin{figure}[htb]
    \centering
    \includegraphics[width=0.70\linewidth]{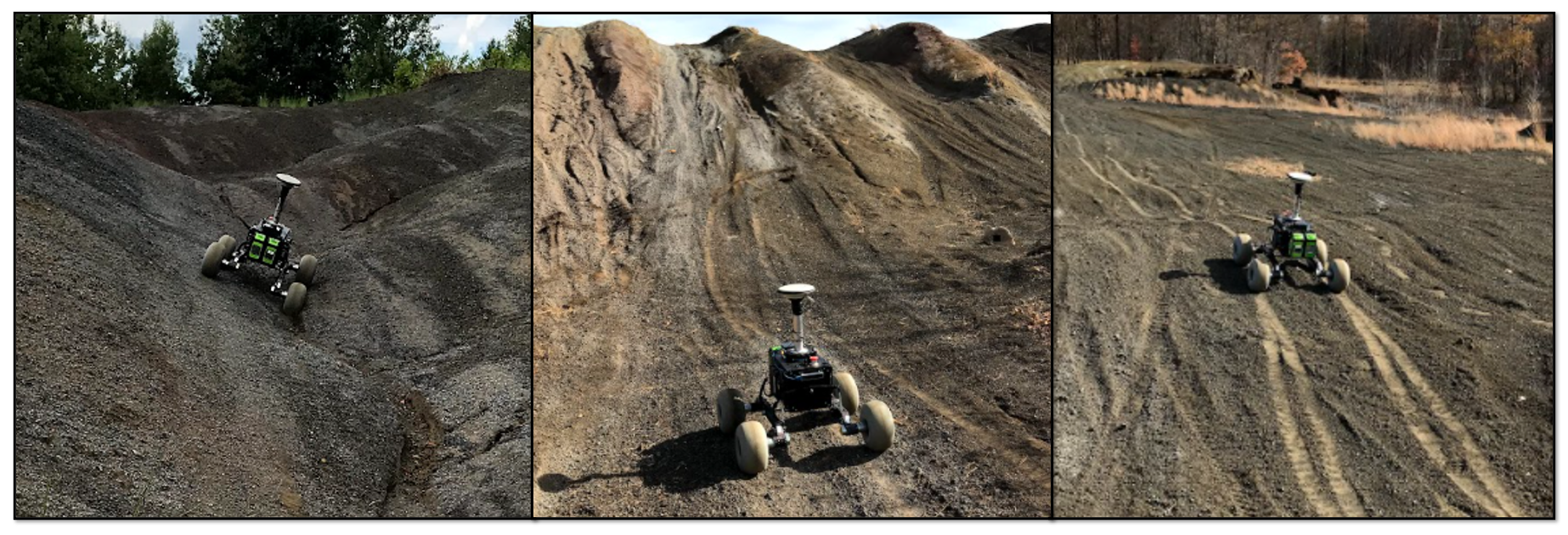}
    \caption{Pathfinder rover during field tests in a Martian-analog terrain located at Point Marion, PA.  }
    \label{fig:pathfinderLast}
\end{figure}

The experiments are performed in a field analog to Martian terrain (see Fig.~\ref{fig:pathfinderLast}). This field consists of burnt-coal ash piles located at Point Marion, PA. The topographic properties of the field include sloped, pitted, fractured, and sandy areas. Moreover, the chemical composition of the field resembles the abundant chemical compounds found in the Martian environment, such as aluminum oxide, iron oxide, silicon dioxide, and calcium oxide~\cite{peters2008mojave, ramme2004we}. For these reasons, both visual and chemical characteristics of the field are considered sufficient for experiments. It is acknowledged that simulating Mars soil requires the measurement of the chemical properties composition by weight composition percent by weight of the soil \cite{ming2017chemical}; however, this is beyond the scope of this study. 
\subsection{Evaluation}

The method is evaluated with several comparison analyses. First, the \ac{VIO} estimates from the tracking system are statistically compared with the proposed method using \ac{RMSE} values of the 3D position estimates. Then, the slip detection accuracy and velocity estimations are examined with respect to the DGPS based velocity and slip detection. Finally, the heading estimation for \ac{WO}, \ac{DGPS}, and proposed method are compared.

\begin{table}[htb]
\centering
\footnotesize
\caption{Statistical Position Error Comparison against VIO}
\label{tab:statVIO}
\centering
\begin{tabular}{@{}lcccccc@{}}
\hline
 & \multicolumn{3}{c}{VIO }   &\multicolumn{3}{c}{Proposed} \\
Case & \multicolumn{3}{c}{RMSE (m)}   &\multicolumn{3}{c}{RMSE (m)} \\
 & \scriptsize{East}& \scriptsize{North}& \scriptsize{Up}& \scriptsize{East}& \scriptsize{North}& \scriptsize{Up} \\
\hline\hline
ShortFast1         &44.04	    &27.24	    &\textbf{0.42}	    &\textbf{1.23}	&\textbf{2.51}	&1.97\\
ShortFast2          &2.70	    &10.41	    &2.12	    &\textbf{1.03}	&\textbf{0.49}	&\textbf{0.65}\\
ShortFast3          &13.27      &56.62	    &\textbf{2.14}	    &\textbf{0.66}	&\textbf{0.40}	&2.89\\
ShortFast4         &\textbf{1.41}	    &12.22	    &1.59       &2.26	&\textbf{1.76}	&\textbf{0.38}\\
ShortSlow           &13.25	    &44.69	    &2.83	    &\textbf{1.01}	&\textbf{0.28}	&\textbf{1.47}\\
\hline
\end{tabular}

\end{table}

\begin{figure}[htb]
\centering
{
	\subfigure[ShortFast1 ]
	{
		\includegraphics[width=0.48\linewidth]{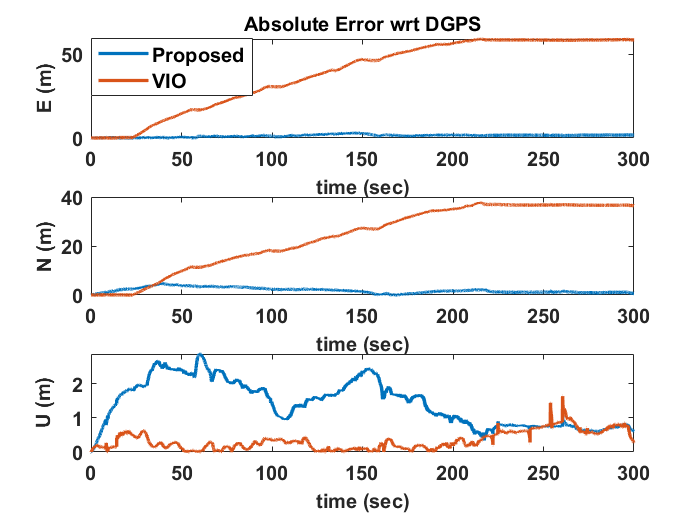}
		\label{fig:fig2SF1}
	}
	\subfigure[ShortFast2]
	{
		\includegraphics[width=0.48\linewidth]{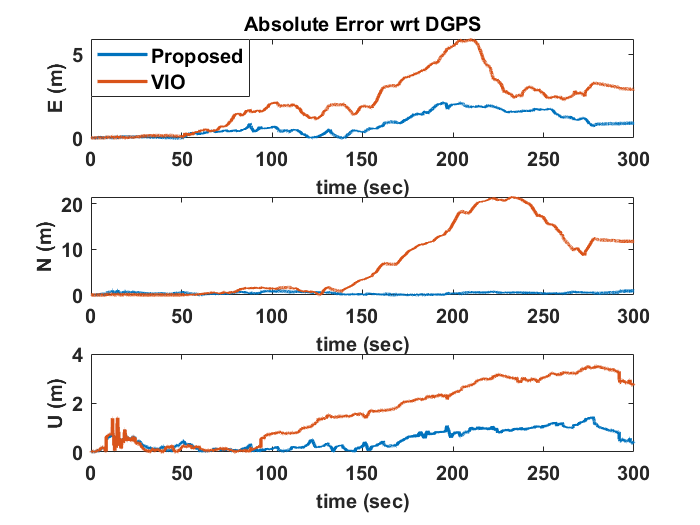}
		\label{fig:fig2SF2}
	}
		\subfigure[ShortFast3]
	{
		\includegraphics[width=0.48\linewidth]{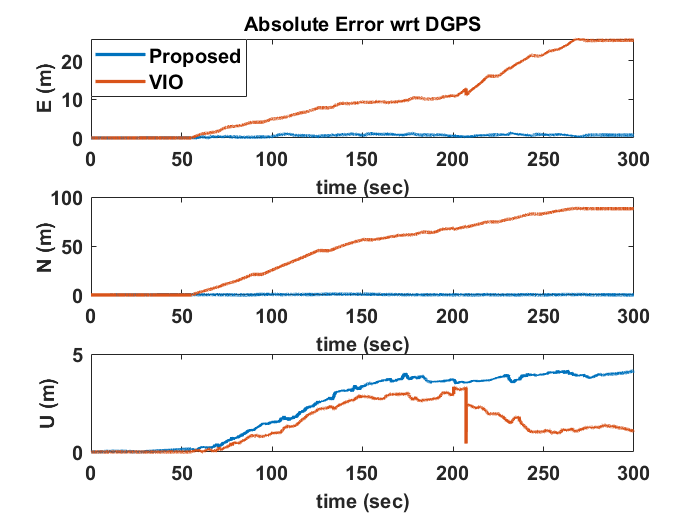}
		\label{fig:fig2SF3}
	}
	\subfigure[ShortSlow]
	{
		\includegraphics[width=0.48\linewidth]{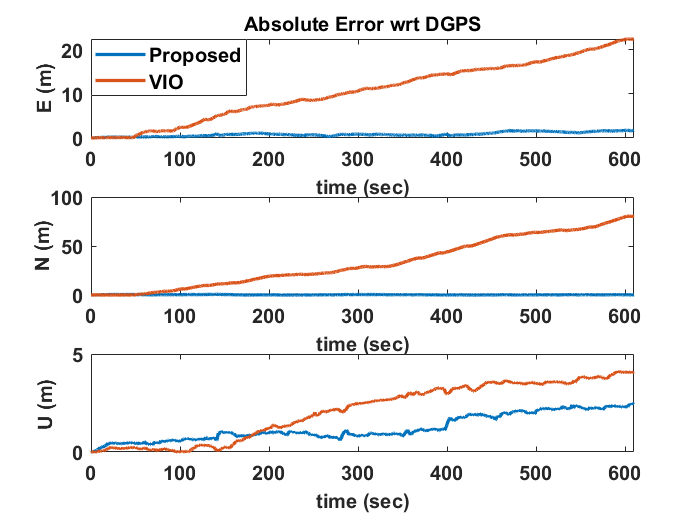}
		\label{fig:fig2SS}
	}

}
\caption{Absolute 3D position (East-North-Up) error accumulation figures with respect to time as the rover travel increases. This figure shows a relationship between the distance traveled and the estimation error of the proposed method and \ac{VIO} estimation. The absolute error is calculated with respect to the \ac{DGPS} solution. }
\label{fig:vio2}
\end{figure}

\begin{figure}[htb]
\centering
{
	\subfigure[ShortFast1 ]
	{
		\includegraphics[trim={8cm 0.5cm 8cm 0.5cm}, width=0.36\textwidth, clip]{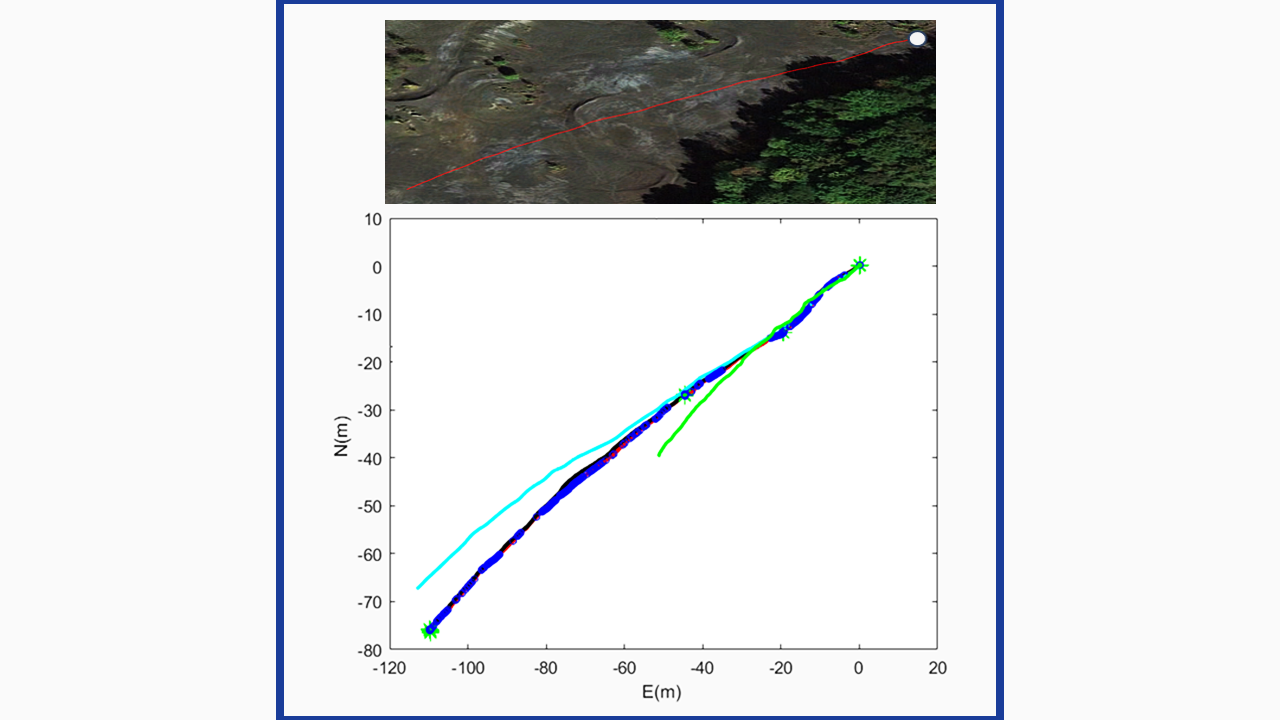}
		\label{fig:fig2}
	}
	\subfigure[ShortFast2]
	{
		\includegraphics[trim={8cm 0.5cm 8cm 0.5cm}, width=0.36\textwidth, clip]{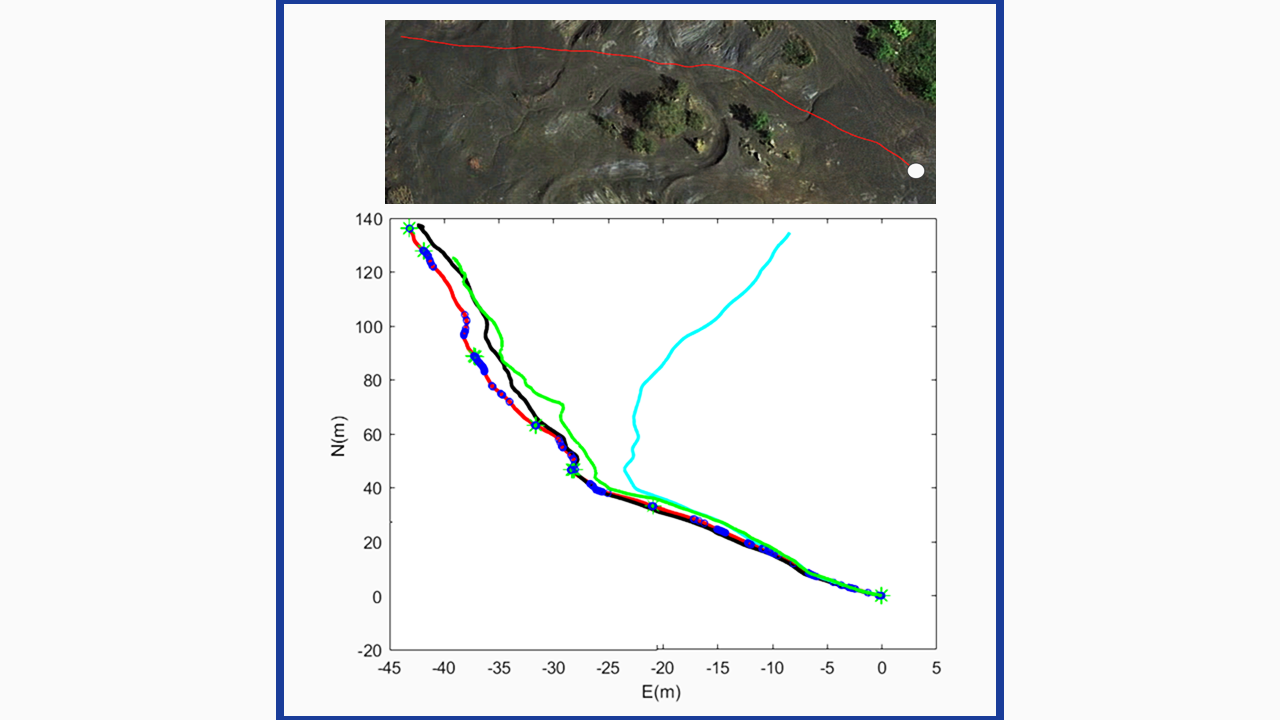}
		\label{fig:fig2}
	}
		\subfigure[ShortFast3]
	{
		\includegraphics[trim={8cm 0.5cm 8cm 0.5cm}, width=0.36\textwidth, clip]{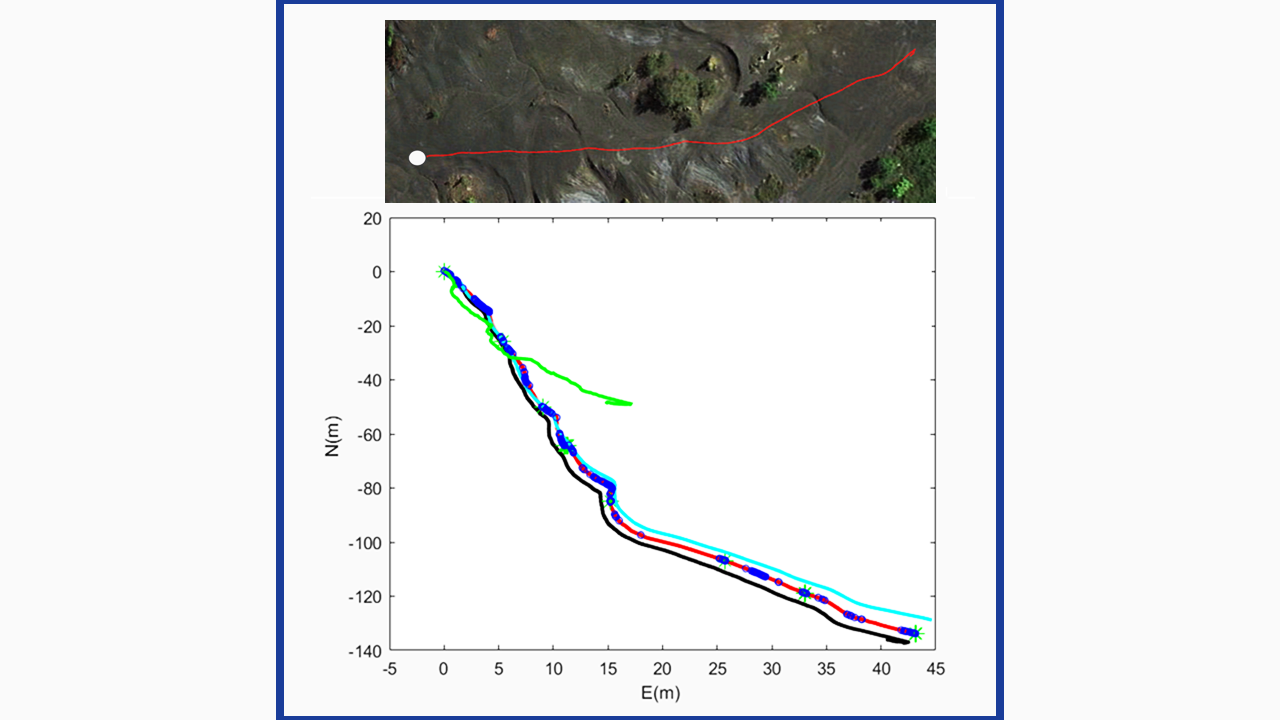}
		\label{fig:fig2}
	}
	\subfigure[ShortSlow]{\includegraphics[trim={10cm 0.5cm 10cm 0.5cm}, width=0.345\textwidth, clip]{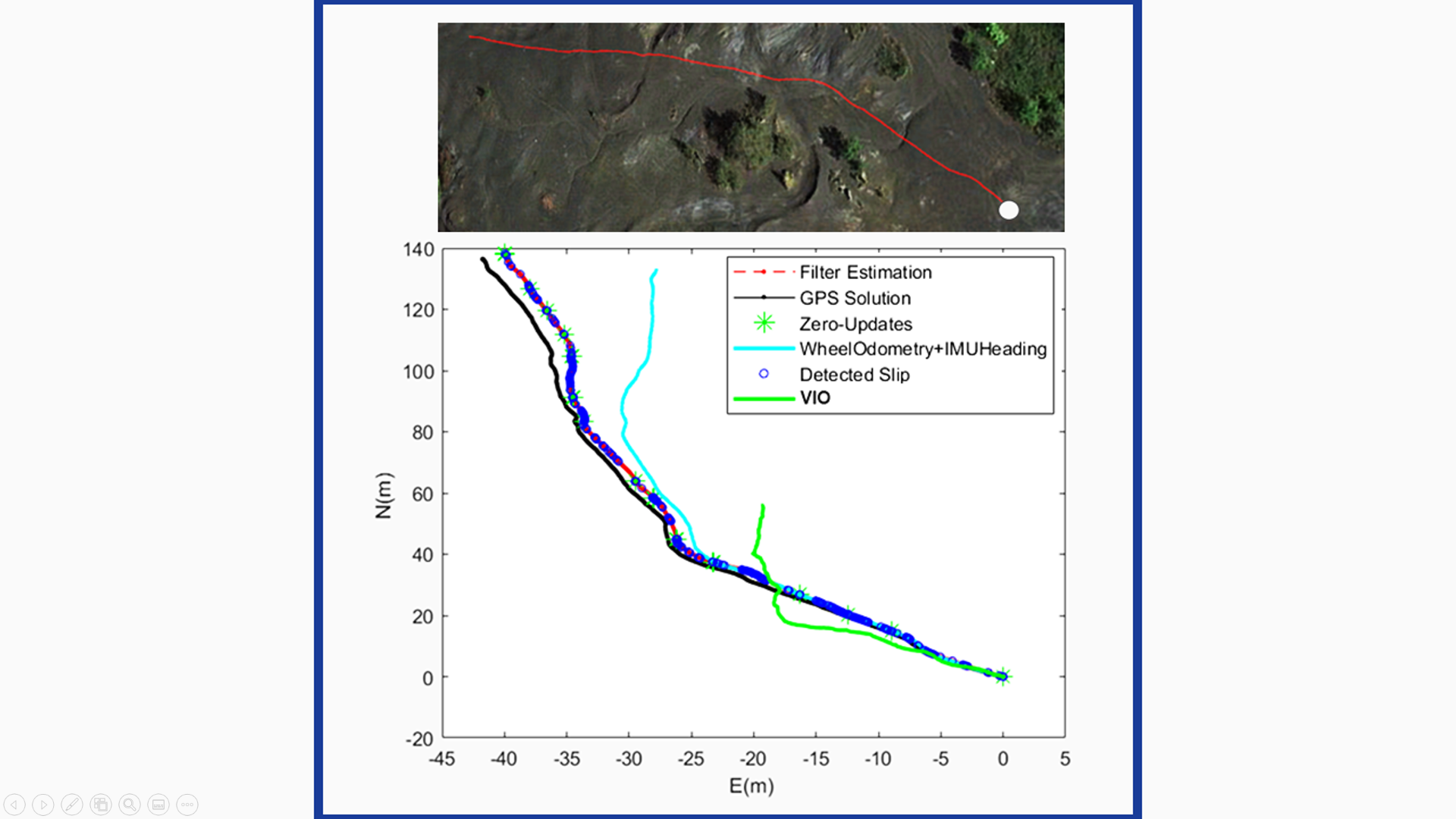}
	\label{fig:purgatory4}
	}

}
\caption{Ground track depictions of the proposed method, \ac{VIO} estimation, \ac{WO} with IMU heading estimation, and \ac{DGPS} solution. The \ac{DGPS} solution is treated as the truth ground track and given as a black line. \ac{VIO} estimation is given as a green line, \ac{WO}-based estimation given as cyan line, and the proposed estimation given as red line with blue circles (the blue circles are the slip values for s$>$ $|$0.2$|$)    }
\label{fig:vio}
\end{figure}

In position estimation comparison field experiment on a perceptually degraded terrain, rover is remotely controlled for approximately 150 m and the comparison results are given for five cases in Table~\ref{tab:statVIO}. The naming convention for the test cases are selected based on relative to the distance traversed and the rover velocity. For example, ShortFast stands for short distance ($\sim$~150 m) with fast speed ($\sim$~0.8 m$/$s), LongFast is long distance ($>$500 m) with fast speed, and ShortSlow is short distance with slow speed ($\sim$~0.3 m$/$s). ShortSlow case is used for two specific observations: 1)~to observe the effect of having a slower velocity in the state estimation, 2)~to observe the INS drift for a longer traversal time. 
Absolute 3D position error accumulation for the ShortFast~1-2-3 and ShortSlow cases in East-North-Up (ENU) frame are given in Fig.~\ref{fig:vio2}.   

\begin{figure}[htb]
\centering
{

	\subfigure[ShortFast4 Ground Track]
	{
		\includegraphics[trim={8cm 0.5cm 8cm 0.5cm}, width=0.42\textwidth, clip]{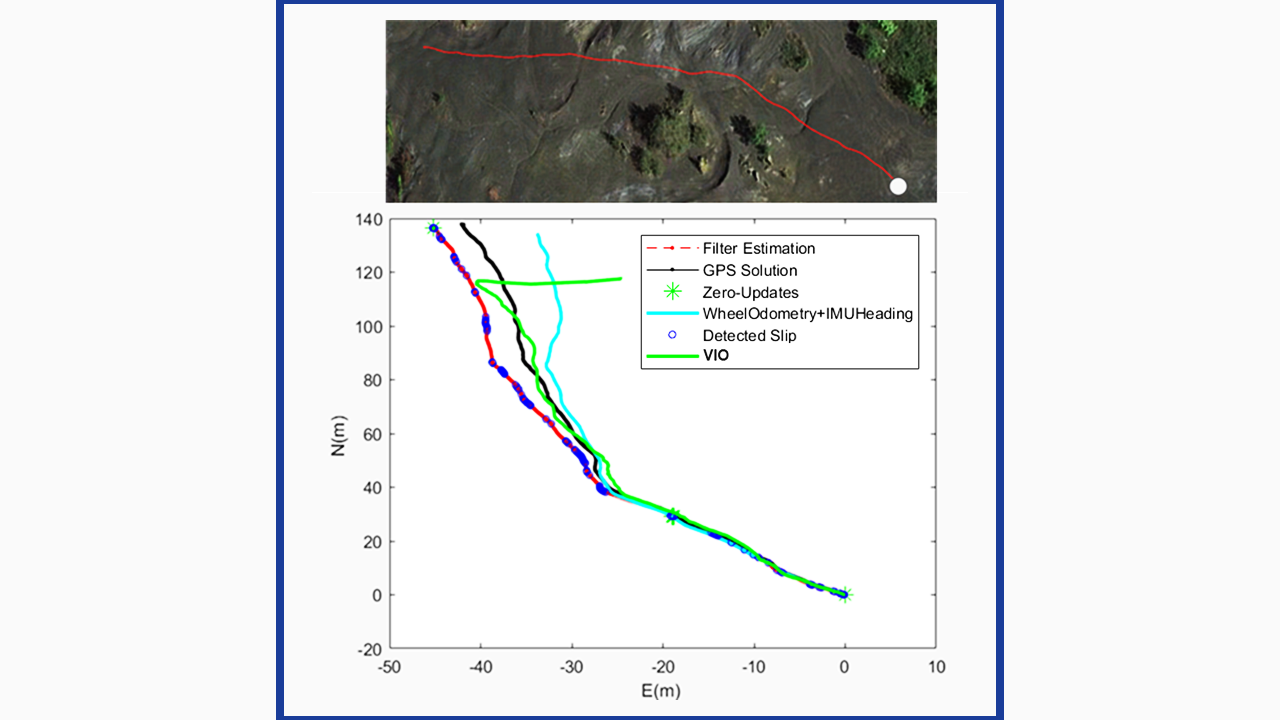}
		\label{fig:fig2}
	}
	\subfigure[ShortFast4 Error Accumulation]
	{
		\includegraphics[width=0.54\linewidth]{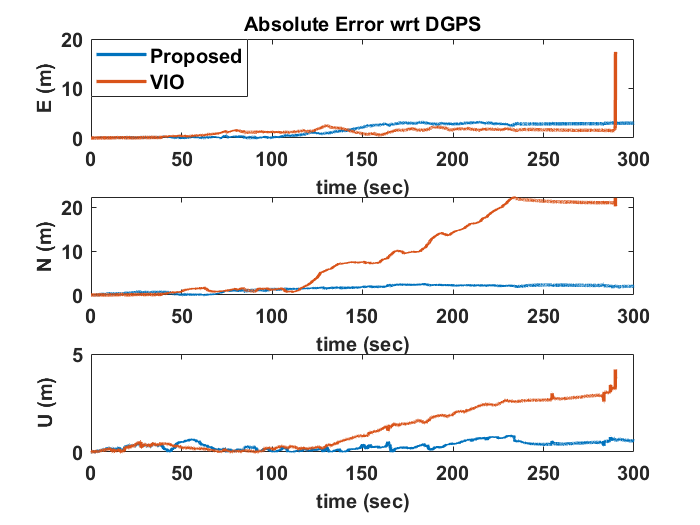}
		\label{fig:fig2SS}
	}
}
\caption{Ground track and error accumulation for ShortFast4 scenario.   }
\label{fig:vio3}
\end{figure}

\begin{figure}[t!]
    \centering
   \subfigure[ShortFast3]{
    \includegraphics[width=0.86\linewidth]{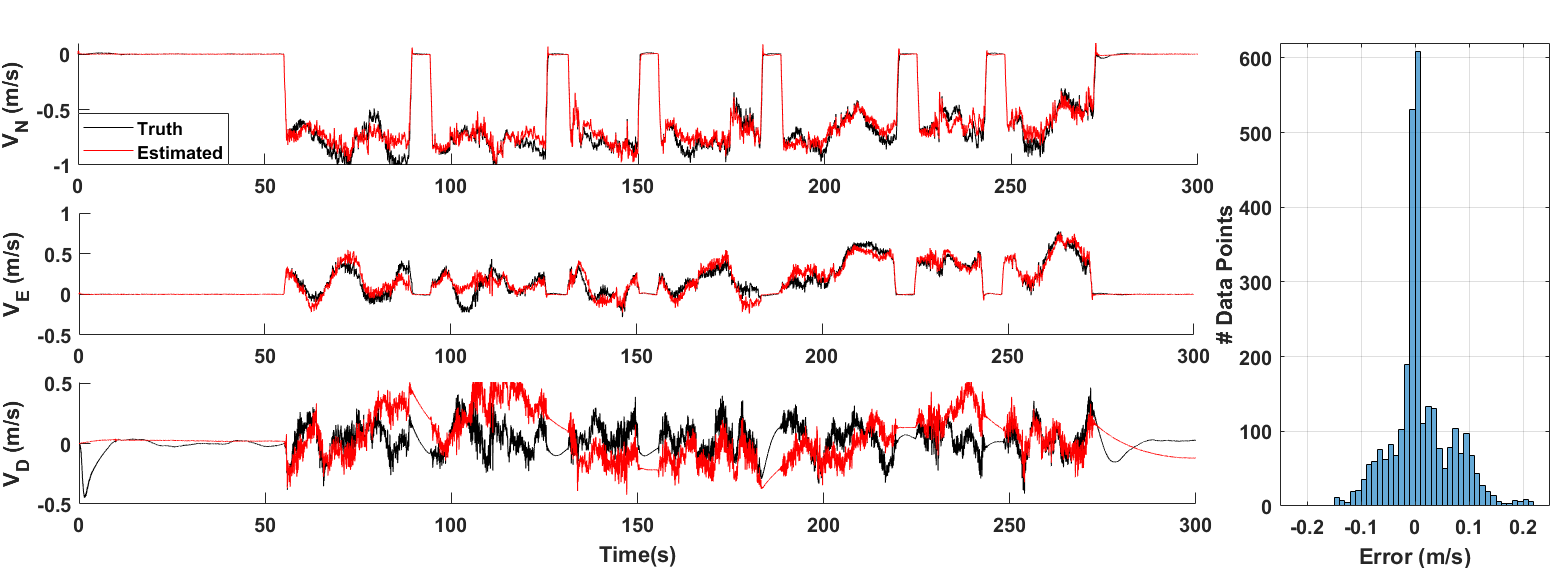}
    \label{fig:ShortFast3}}

    \subfigure[ShortSlow]{
   \includegraphics[width=0.86\linewidth]{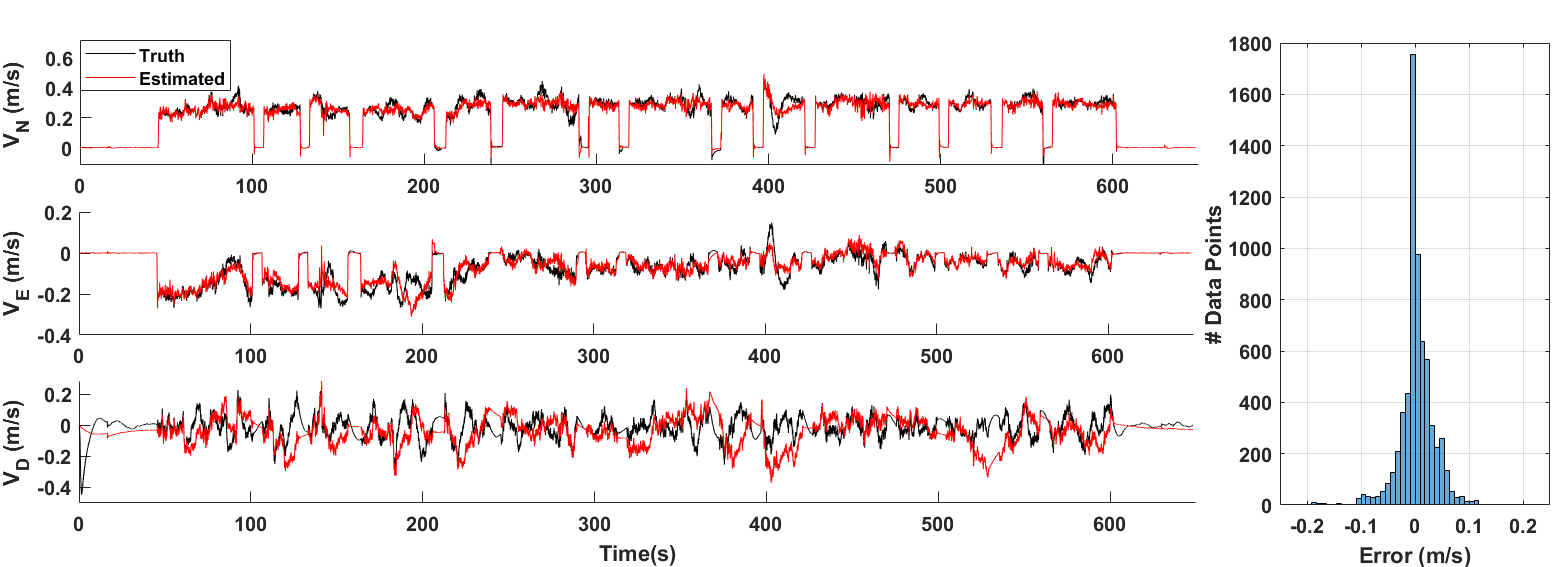}
    \label{fig:ShortSlow}}
    
    \subfigure[LongFast]{
   \includegraphics[width=0.86\linewidth]{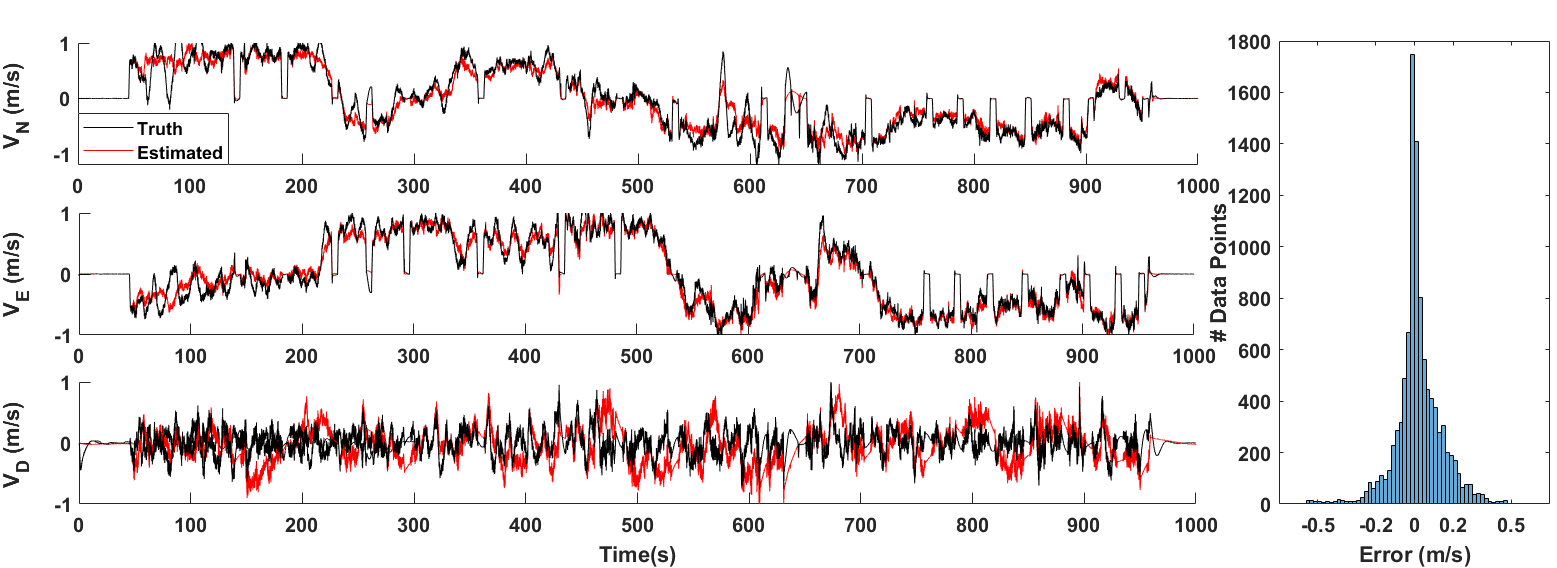}
    \label{fig:LongFast}}
    
  \caption{Left figures show the velocity estimation compared to post processed DGPS velocity solution that assumed as truth velocity both in North, East, Down (NED) frame. Right histogram figures are the velocity error distributions between the estimated and the truth velocities. }
  \label{fig:velocities}
\end{figure}

In areas with many visual features, the tracking system can produce dependable solutions, but it often fails in places with few detectable and trackable elements. This is a typical problem with visual-based localization methods since they require a decent number of distinct visual components to function properly.~\cite{campos2020orb,strader2020perception}. For example, it is shown that the \ac{VO} method working on Curiosity rover has a remarkable convergence rate in one of the most recent works~\cite{rankin2020driving}; however, in the same work, it can be seen that the majority of the VO failures on MSL are due to scarcity of sufficient features at sandy terrains. In our analysis in a visually low-feature environment, we observed the similar issues for using VO which can be seen in Fig.~\ref{fig:vio}. In addition to that, in ShortFast4 test, the VIO failed after 124~m of traversal (see Fig.~\ref{fig:vio3}). Moreover, even it can keep the East positioning in a reasonable accuracy for ShortFast2 and ShortFast4 tests, the position estimations of the North axis are more than 10 m RMSE for all tests. On the other hand, the proposed method in this work often outperforms the VIO solution in this environment. The estimated velocity is compared with the post processed DGPS-based velocity solution to evaluate the performance of the proposed velocity estimation model and the overall distribution of the translational velocity errors in the tests are given in Fig.~\ref{fig:velocities}.

Based on both previous literature and Curiosity rover thresholds for slip~\cite{arvidson2017mars, skonieczny2019data}, the measured absolute slip ratio is classified in five ranges; 1) No Slip ($s\approx0$), 2) Low Slip ($0<s<=0.2$), 3) Medium Slip ($0.2<s<=0.4$), 4) High Slip ($0.4<s<=0.7$), and 5) Extreme Slip ($0.7<s<=1.0$). Using these slip ratio ranges, confusion matrices of estimated slip values versus truth (DGPS) slip values are quantitatively compared in Table~\ref{tab:slipacc} for LongFast (LF), ShortSlow (SS), ShortFast2 (SF2), and ShortFast3 (SF3) scenarios. In order to qualitatively represent the regions for the detected slip ratio values, classification figures with the absolute slip ratio data points are given in Fig.~\ref{fig:vio2} for the estimated slip values with respect to the truth. The slip detection is performed each time when the wheel encoder data is available (10~Hz). The truth slip detection rate is also 10~Hz. The estimated slip detection accuracy is calculated using DGPS-based slip detection such that 

\begin{equation}
\label{slip}
s_{TR}=\left\{\begin{array}{lll}
{1-\frac{{v_x}_{TR}}{r \omega}} & { \text { (if }\omega \neq 0, {v_x}_{TR}<r \omega,  \text { s $>$ 0 })} \\
{\frac{r \omega}{{v_x}_{TR}}-1} & { \text { (if }{v_x}_{TR} \neq 0, {v_x}_{TR}>r \omega,  \text { s $<$ 0 })} \\
0 & { \text { (if } {v_x}_{TR} = r\omega \lor {v_x}_{TR} = \omega = 0  )}
\end{array}\right., \quad s_{TR} \in[-1,1] 
\end{equation}

where $s_{TR}$ is DPGS-based slip ratio.

\begin{table} [b!]
\centering
\footnotesize
\caption{Confusion matrices of the estimated slip ratio values versus measured DGPS-based slip ratio classification. LF is LongFast (659m, 0.8m$/$s), SS is ShortSlow (146m, 0.3m$/$s), SF2 is ShortFast2 (145m, 0.8m$/$s), and SF3 is ShortFast3 (148m, 0.8m$/$s). The diagonal values of the column normalized matrix corresponds to the cases where the estimation identifies the same class as the truth. Each entry represents the percentage of DGPS-based slip ratio that is estimated by the slip detector. E.g., in LF, 7.1$\%$ of low slip classified as medium slip, 2.6$\%$ of medium slip classified as high slip. }  
\label{tab:slipacc}
\centering
\begin{tabular}{@{}clccccc|lccccc@{}}
& &\multicolumn{5}{c}{Truth} & & \multicolumn{5}{c}{Truth}\\
\hline
&LF \vline &\multicolumn{1}{c}{NoSlip}&\multicolumn{1}{c}{Low}& \multicolumn{1}{c}{Med}   &\multicolumn{1}{c}{High} &\multicolumn{1}{c}{Ext}  \vline & SS \vline &\multicolumn{1}{c}{NoSlip}&\multicolumn{1}{c}{Low}& \multicolumn{1}{c}{Med}   &\multicolumn{1}{c}{High} &\multicolumn{1}{c}{Ext} \\
\hline
\multirow{5}{*}{\rotatebox[origin]{90}{{\centering Estimated}}}&NoSlip  & \textcolor{white}{98.3$\%$} \cellcolor[gray]{.1}	&0.7$\%$	&0.4$\%$	&0.3$\%$	&0.4$\%$	
&NoSlip  &\textcolor{white}{97.5$\%$} \cellcolor[gray]{.1}	&1.0$\%$	&0.2$\%$	&0.0$\%$	&0.0$\%$\\
&Low     &1.4$\%$	&\textcolor{white}{91.9$\%$} \cellcolor[gray]{.1}	&77.1$\%$\cellcolor[gray]{0.3}	&42.2$\%$\cellcolor[gray]{0.6}	&39.0$\%$ \cellcolor[gray]{.7}
&Low     &2.4$\%$	&\textcolor{white}{96.0$\%$} \cellcolor[gray]{.1}	&81.3$\%$\cellcolor[gray]{.2}	&39.6$\%$\cellcolor[gray]{.6}	&0.0$\%$\\
&Med     &0.2$\%$	&7.1$\%$	&\textcolor{white}{19.8$\%$}\cellcolor[gray]{.8}	&30.2$\%$\cellcolor[gray]{.7}	&8.5$\%$
&Med     &0.1$\%$	&3.0$\%$	&\textcolor{white}{17.4$\%$}\cellcolor[gray]{.8}	&34.0$\%$\cellcolor[gray]{.7}	&0.0$\%$\\
&High    &0.1$\%$	&0.3$\%$	&2.6$\%$	&\textcolor{white}{24.0$\%$}\cellcolor[gray]{.8}	&13.6$\%$
&High    &0.0$\%$	&0.0$\%$	&0.9$\%$	&\textcolor{white}{22.6$\%$}\cellcolor[gray]{.8}	&5.0$\%$\\
&Ext     &0.0$\%$	&0.1$\%$	&0.2$\%$	&3.2$\%$	&\textcolor{white}{38.6$\%$}\cellcolor[gray]{.6}
&Ext     &0.0$\%$	&0.0$\%$	&0.2$\%$	&3.8$\%$	&\textcolor{white}{95.0$\%$}\cellcolor[gray]{.1}\\
\hline
&$\#$  Data     &2303	&5627	&1809	&308	&236
&$\#$  Data     &1778	&4049	&563	&53	&40\\
\hline
\hline
&SF2 \vline &\multicolumn{1}{c}{NoSlip}&\multicolumn{1}{c}{Low}& \multicolumn{1}{c}{Med}   &\multicolumn{1}{c}{High} &\multicolumn{1}{c}{Ext}  \vline & SF3 \vline &\multicolumn{1}{c}{NoSlip}&\multicolumn{1}{c}{Low}& \multicolumn{1}{c}{Med}   &\multicolumn{1}{c}{High} &\multicolumn{1}{c}{Ext} \\
\hline
\multirow{5}{*}{\rotatebox[origin]{90}{{\centering Estimated}}}&NoSlip  &\textcolor{white}{99.5$\%$}\cellcolor[gray]{.1}	&0.6$\%$	&0.3$\%$     &0.0$\%$	&0.0$\%$ 
&NoSlip  &\textcolor{white}{99.1$\%$}\cellcolor[gray]{.1}	&0.9$\%$	&0.0$\%$	&0.0$\%$	&0.0$\%$\\
&Low     &0.5$\%$	&\textcolor{white}{95.5$\%$}\cellcolor[gray]{.1}	&84.4$\%$\cellcolor[gray]{.2}	&29.4$\%$\cellcolor[gray]{.7}	&0.0$\%$ 
&Low     &0.9$\%$	&\textcolor{white}{96.0$\%$}\cellcolor[gray]{.1}	&64.5$\%$\cellcolor[gray]{.4}	&0.0$\%$	&0.0$\%$\\
&Med     &0.0$\%$	&3.8$\%$	&\textcolor{white}{14.0$\%$}\cellcolor[gray]{.8}	&29.4$\%$\cellcolor[gray]{.7}	&0.0$\%$ 
&Med     &0.0$\%$	&3.0$\%$	&\textcolor{white}{32.6$\%$}\cellcolor[gray]{.7}	&42.1$\%$\cellcolor[gray]{.6}	&0.0$\%$\\
&High    &0.0$\%$	&0.1$\%$	&1.4$\%$	&\textcolor{white}{35.3$\%$}\cellcolor[gray]{.7}	&14.3$\%$ 
&High    &0.0$\%$ 	&0.1$\%$  &2.3$\%$ 	&\textcolor{white}{36.8$\%$} \cellcolor[gray]{.7}	&0.0$\%$ \\
&Ext     &0.0$\%$	&0.0$\%$	&0.0$\%$	&5.9$\%$	&\textcolor{white}{85.7$\%$}\cellcolor[gray]{.2} 
&Ext     &0.0$\%$	&0.0$\%$	&0.7$\%$	&21.1$\%$\cellcolor[gray]{.8}	&\textcolor{white}{100.0$\%$}\cellcolor[gray]{.1}\\
\hline
&$\#$ Data     &1263	&1543	&365	&34	&14
&$\#$  Data     &1133	&1534	&307	&19	&5\\
\hline
\end{tabular}
\end{table}

\begin{figure}[htb]
\centering
{
	\subfigure[LongFast ]
	{
		\includegraphics[width=0.46\linewidth]{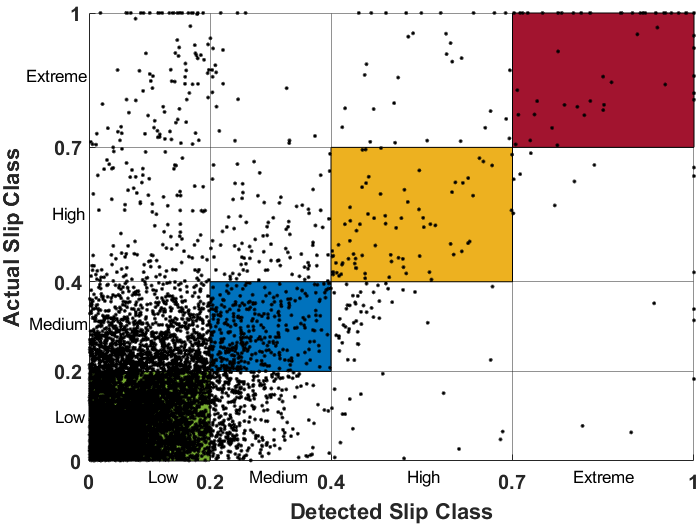}
		\label{fig:longfast_slip}
	}
	\subfigure[ShortSlow]
	{
		\includegraphics[width=0.46\linewidth]{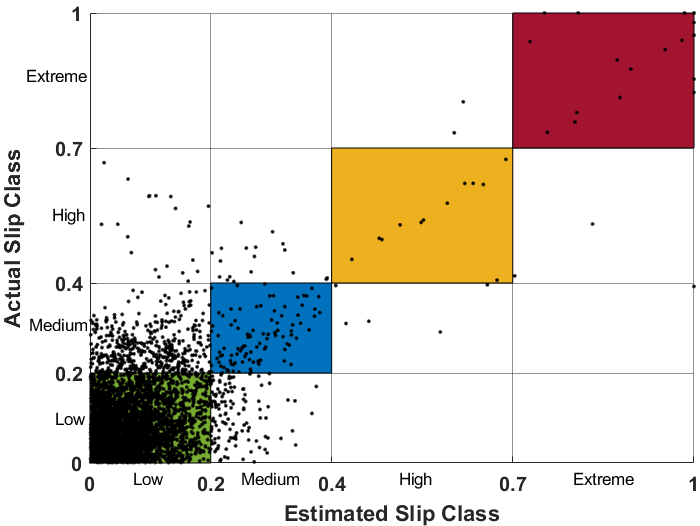}
		\label{fig:shortslow_slip}
	}
		\subfigure[ShortFast2]
	{
		\includegraphics[width=0.46\linewidth]{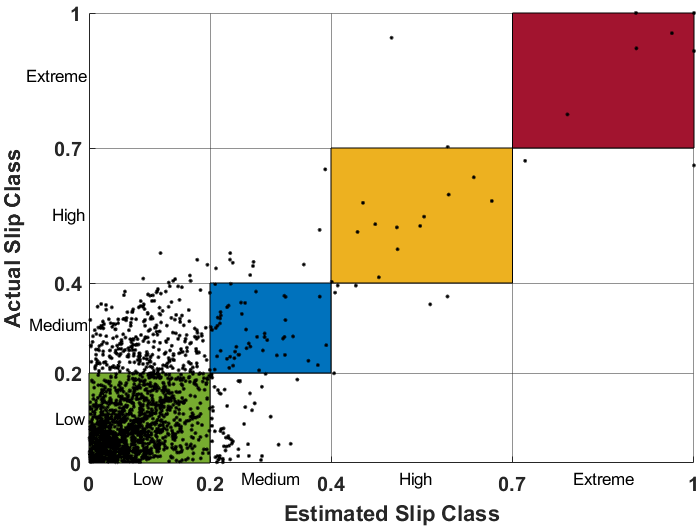}
		\label{fig:shortfast2_slip}
	}
	\subfigure[ShortFast3]
	{
		\includegraphics[width=0.46\linewidth]{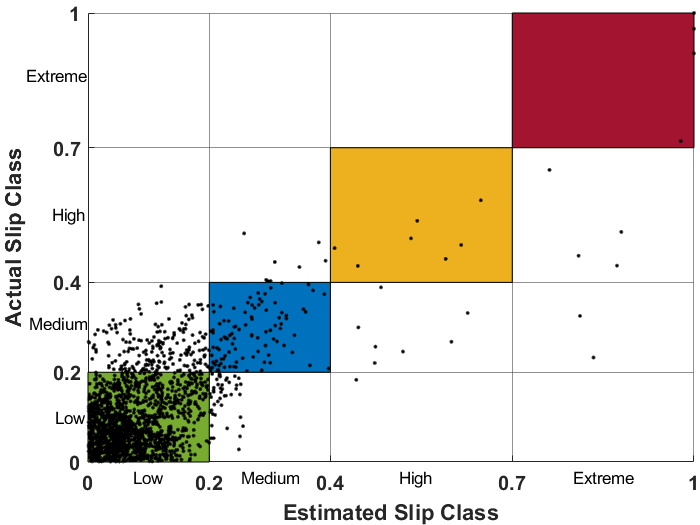}
		\label{fig:shortfast3_slip}
	}

}
\caption{Depictions of the detected absolute slip ratio data in the slip classification. The black data points in each column represents the estimated class of detected slip and the data points in each row represent their actual (DGPS-based) slip class. For example, the data points in the area between 0.2 - 0.4 range (Medium) Estimated Slip Class and 0.4 - 0.7 range (High) Actual Slip Class indicate the detected high slip values are estimated as medium slip. The data points in colored regions indicate when the detected slip point and the actual slip are classified as the same. }    
\label{fig:vio11}
\end{figure}

In these test cases, the LongFast (LF) has the longest traversal distance ($\sim$ 650 m), which is an additional test case to show the slip detection accuracy limit for longer distances, whereas other test cases are between 140 m to 150 m.  The confusion matrices in Table~\ref{tab:slipacc} shows the proprioceptive slip detection accuracy for different ranges of slip values. In addition to the detection, the slip range classification performed reliably for no-slip and low slip ranges for all cases, and extreme slip is identified well for short distances. Most of the medium slip values were classified as low slip. Similar to the medium slip range, in general, high slip class overlaps between medium and low slip. Even the slip detection performed well, this class overlap may indicate the fuzzy boundaries of the slip regions for classification. According to an experimental work on wheel-soil interaction characteristics in~\cite{shirai2015development}, such a boundary is mainly due to the soil bulldozing and transportation effect. In order to visualize these ambiguous boundaries, the values of the detected slip data in Table~\ref{tab:slipacc} are plotted and given in Fig.~\ref{fig:vio11}. 

The accuracy decrease for longer distances (e.g., more than 500 m) could be attributed to the limitation of using a dead-reckoning method without any external updates and also the quality of the \ac{IMU}. The \ac{IMU} used in this study is a relatively low-cost sensor compared to Northrop Grumman LN-200S IMU with fiber optic gyroscopes and solid state silicon MEMS acceloremeters used in MERs, Curiosity, and Perseverance. The error histograms in Fig.~\ref{fig:velocities} can also be interpreted as the accuracy of the slip detection. Overall, these cases support the reliability of the method for distances around 150 m in the test field for the used rover platform with the given sensor setup, and also reveal the limitation of the method for the longer distances without external update.

\begin{figure}[t!]
\centering
{
\subfigure[LongFast ]{
		\includegraphics[width=0.8\columnwidth]{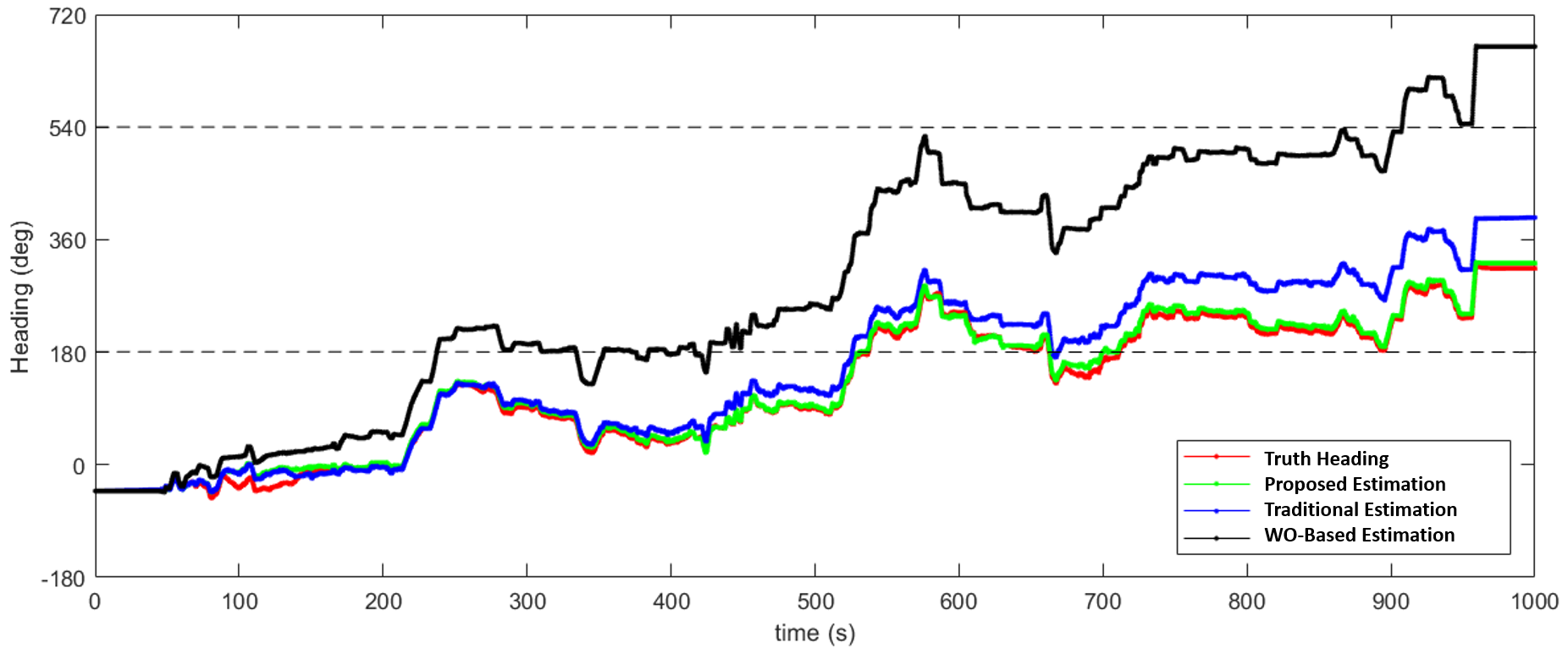}
		\label{fig:purgatory3}}
\subfigure[ShortSlow]{
		\includegraphics[width=0.48\columnwidth]{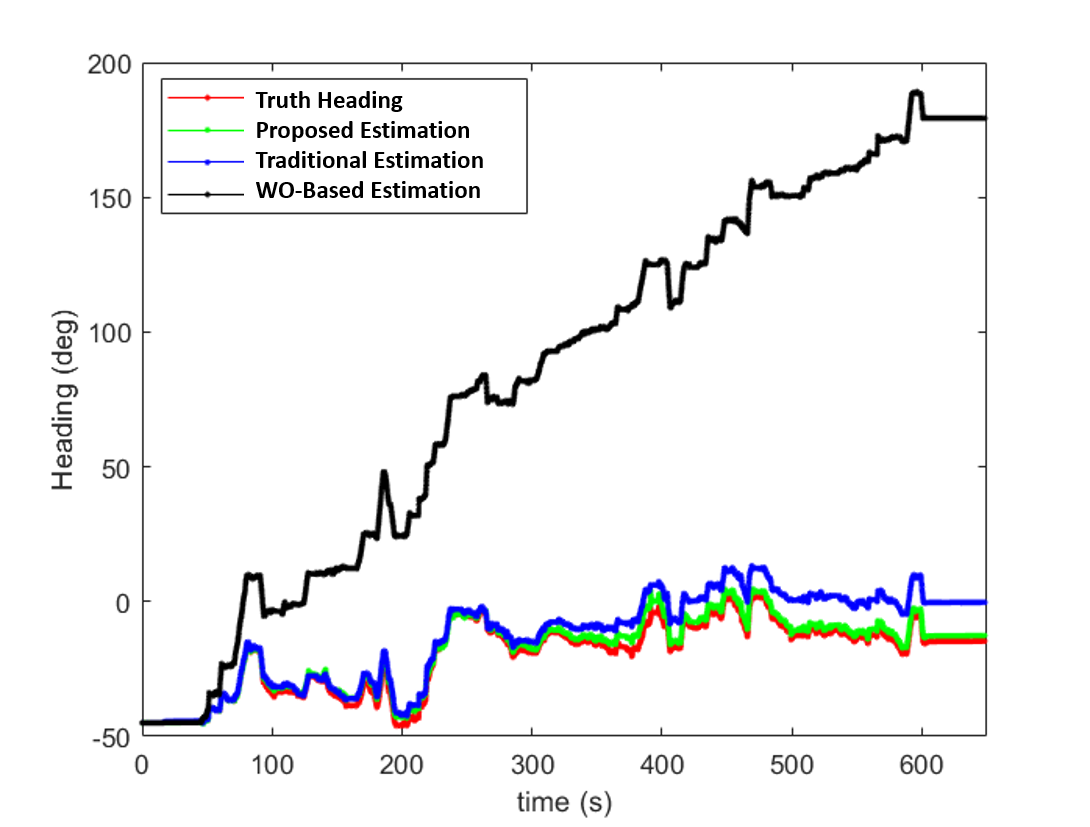}
		\label{fig:fig2}}
\subfigure[ShortFast2]{
        \includegraphics[width=0.48\columnwidth]{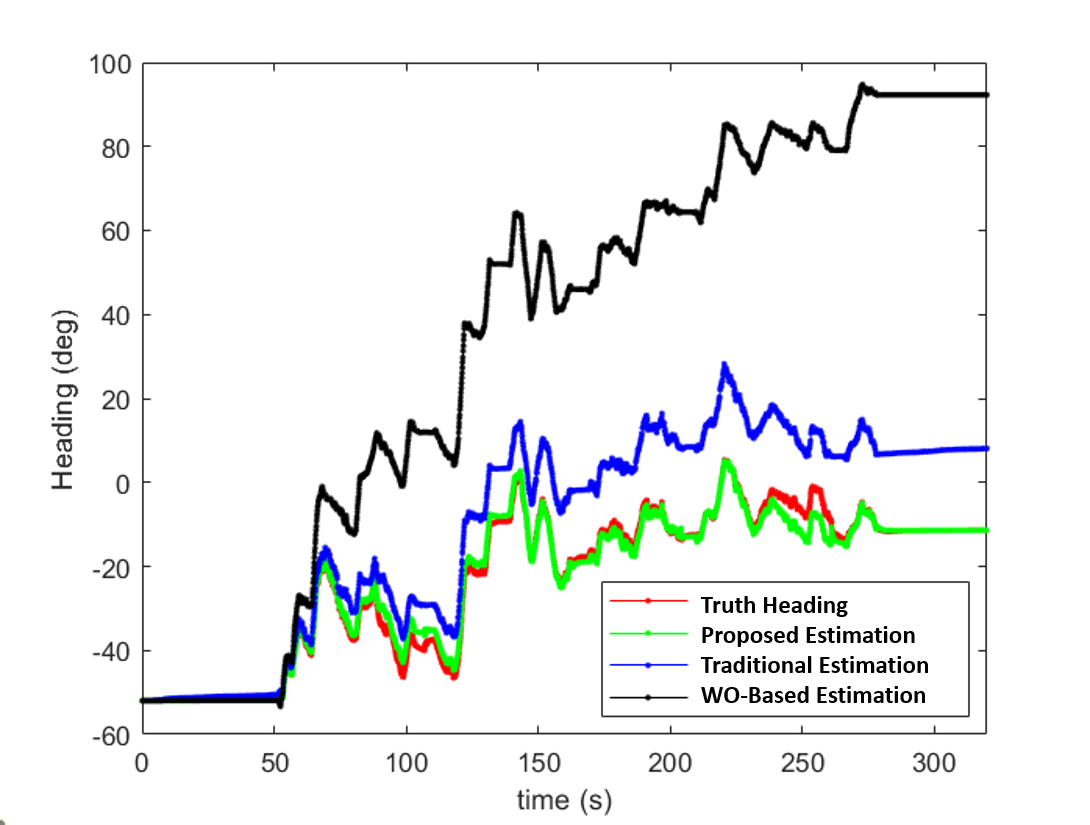}
		\label{fig:fig2}}		
}
\caption{The unwrapped heading estimation comparisons. The heading estimation accuracy is critically important to have a reliable translational velocity for the slip detection. Proposed method overall accuracy at the end amount, LongFast: 8~$\deg$, ShortSlow: 2~$\deg$, ShortFast = 0.01~$\deg$. Traditional direct integration estimation at the end amount, LongFast: 74~$\deg$, ShortSlow: 14~$\deg$, ShortFast: 17~$\deg$.}
\label{fig:heading}
\end{figure}

The slip detection in this work depends on the velocity estimation accuracy. Specifically, the translational velocity is used for calculating the longitudinal slip ratio.  Given that the attitude of the robot is used to integrate the velocity vector in the INS mechanism, the attitude estimation accuracy is critically important to have a reliable translational velocity for the slip detection. In this respect, to further evaluate the method, the estimated heading with the proposed architecture, directly integrated heading estimation, and wheel encoder based heading are compared with the DGPS-based heading estimation in Fig.~\ref{fig:heading}. 

In Fig.~\ref{fig:heading}, the unwrapped heading comparisons show the error growth of unbounded estimations for three different driving time interval including 1000 s (long), 600 s (medium), and 300 s (short). After a relatively short drive, \ac{WO}-based heading estimation accumulates significant heading error which is a well-known problem due to wheel slippage. Also considering the used rover is a skid-steer rover, any small maneuvers during traversal adds more error to the \ac{WO}-based heading estimation. Using a direct integration from the IMU for heading estimation could be useful for short drives. This heading estimation is usually used to support the wheel odometry such that using the velocities from the wheel encoders and using the heading information from the \ac{IMU}. However, when the rover drives longer time, the drift becomes more significant. This is also related to the grade of the used \ac{IMU}, such that given a higher grade \ac{IMU} may provide better estimations for longer drives without any external update (e.g., sun sensors). On the other hand, the proposed method, with using pseudo-measurements, closely follows the truth (DGPS) heading. The overall accuracy at the end amount for LongFast test is 8~$\deg$ (Fig.~\ref{fig:heading}(a)), for ShortSlow is 2~$\deg$ (Fig.~\ref{fig:heading}(b)), and for ShortFast is 0.01~$\deg$ (Fig.~\ref{fig:heading}(c)) for these test cases. The comparison shows that the proposed method outperforms the traditional heading estimation techniques, which can be leveraged to estimate heading for short and medium driving duration reliably; however, it requires external updates after a long driving time.


\section{Conclusion and Future Work} \label{sec:conclusion}
In this work, we employ a method for slip estimation for the cases when visual based perception is not available. Slip estimation is only performed by VO-based methods in current rover operations; however, a proprioceptive slip estimation technique can provide complementary information to current and future missions in the perceptually degraded environments. Since the proposed method uses an INS-based state estimation method, the velocity estimations must be reliable to accurately detect the wheel slippage. Without external aiding, inertial  sensor-based state estimations inherently exhibit accumulated  error. This error accumulation can be alleviated with pseudo-measurements in certain conditions during the traversal. Using pseudo-measurements in a proprioceptive localization method do not require any specific sensor observations rather than the acquisition of IMU and wheel encoder data. In this respect, they provide a cost-effective solution to compensate the INS drift. The effectiveness of the proposed method is demonstrated in a perceptually degraded planetary-analog environment by qualitatively comparing with commercially off the shelf \ac{VIO} solution, wheel encoder based velocity estimation, and \ac{DGPS} velocity solutions. Following the same name convention as is used in this study, the datasets used are made publicly available. 

There are also limitations and several assumptions in this work. In this regard, potential improvements and future work are provided.
Pathfinder has deformable polyurethane slick wheels. Due to physical characteristics of the used wheels, the rover prone to encounter more slippage. This helps to detect slippage with larger frequency and occurrence; however, it also degrades the localization performance significantly. Using slick wheels limits to test the algorithm on unconsolidated soils with high slope values (more than 30 degree). These wheels are non-representative wheels for planetary missions because maximizing wheel traction is one of the most important design criteria for the rovers. Using the algorithms with better representative wheels such as mesh-woven spring wheels or aluminum made wheel design choices will be our next research direction. Another limitation is that the used \ac{IMU} in this study is a relatively low-cost MEMS device with limited sensing quality comparing to what is used in planetary rovers.     
Since the blind driving is a dead-reckoning technique (which causes the uncertainty of the state of the rover to increase with distance), this technique can only be used over short distances in most situations. The length is limited to the distance chosen as safe-to-drive by the rover planners (e.g., 25 m - 60 m), which is based on the rover camera visibility range, prior to employing this mode. We tested our localization algorithm with a maximum 670 m length of blind driving to see its operational limits. Rover safety is more important than accuracy for planetary missions, therefore, these lengths of drivings without human-in-the-loop process may not be suitable for planetary rovers in a manner of rover safety. In fact, most of the short-range tests ($\sim$~150~m) in our tests are longer than the Curiosity rover's longest drive (142.5 m) by sol 2488~\cite{rankin2020driving}.  

\subsubsection*{Acknowledgments}
This research was supported in part by NASA EPSCoR Research Cooperative Agreement WV-80NSSC17M0053, and the Benjamin M. Statler Fellowship.  

\bibliographystyle{apalike}
\bibliography{references}

\end{document}